\documentclass{article}
\usepackage{iclr2027_conference,times}

\usepackage{amsmath,amsfonts,bm}
\usepackage{mathtools}

\DeclarePairedDelimiterX{\cond}[2]{(}{)}{#1 \;\delimsize\vert\; #2}
\newcommand{\given}{\,|\,}

\def\eqref#1{equation~\ref{#1}}

\def\plaineqref#1{\ref{#1}}

\def\1{\bm{1}}

\DeclareMathAlphabet{\mathsfit}{\encodingdefault}{\sfdefault}{m}{sl}
\SetMathAlphabet{\mathsfit}{bold}{\encodingdefault}{\sfdefault}{bx}{n}

\def\sP{{\mathbb{P}}}

\definecolor{ctrlblue}{RGB}{238,244,251}

\newcommand{\skill}{\mathcal{Z}}                        %

\usepackage{amsmath, amsfonts, amssymb, amsthm}
\newtheorem{theorem}{Theorem}[section]

\newtheorem{proposition}[theorem]{Proposition}
\newtheorem{definition}[theorem]{Definition}

\usepackage{url}
\usepackage{graphicx}
\usepackage{booktabs}
\usepackage{multirow}
\usepackage{tabularx}
\usepackage[table]{xcolor}
\usepackage{siunitx}
\usepackage{amsmath}
\usepackage{amssymb}
\usepackage{soul}
\usepackage{wrapfig}
\usepackage{enumitem}

\usepackage[colorlinks=true,
    linkcolor=linkblue,
    citecolor=linkblue,
    urlcolor=linkblue]{hyperref}

\definecolor{linkblue}{RGB}{38, 99, 145}
\definecolor{ctrlblue}{RGB}{238,244,251}

\title{Learning Expressive and Compositional Motion Representation via Spectral Skills}

\author{
Feiyang Wu$^{1,*}$,
Chenxiao Gao$^{1,*}$,
Chen Yang$^{1}$,
Ye Zhao$^{1}$,
Bo Dai$^{1,\dagger}$,
Anqi Wu$^{1,\dagger}$
\\
$^{1}$Georgia Institute of Technology
\\
\texttt{\{feiyangwu, cgao, cyang711, yezhao, anqiwu\}@gatech.edu}
\\[1mm]
{\small $^{*}$Equal contribution.
\qquad
$^{\dagger}$Equal advising.}
}

\iclrfinalcopy %
\begin{document}

\maketitle

\fancyhead{}
\fancyhead[L]{Preprint}

\begin{figure}[!h]
\vspace{-0cm}
    \centering
    \includegraphics[width=1.0\linewidth]{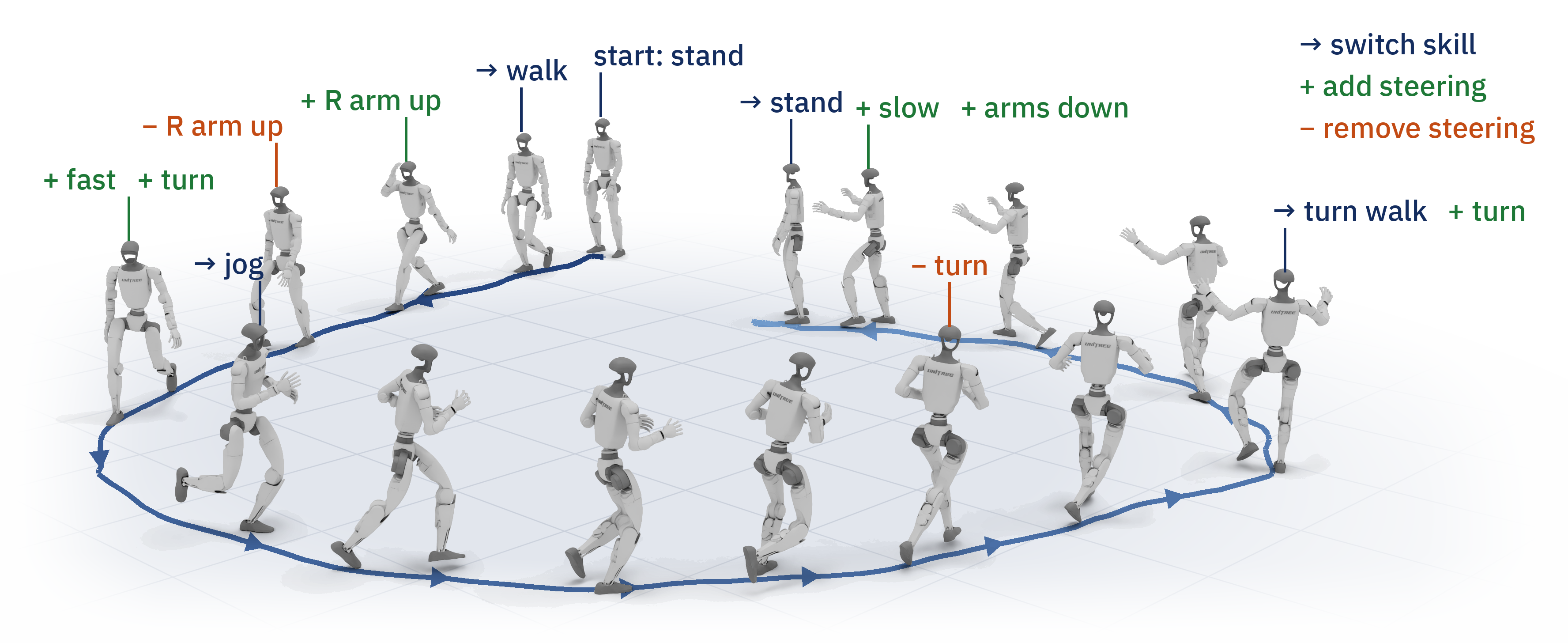}
    \caption{\textbf{Skill chaining and composition with one controller.} One frozen controller chains skills and adds or removes steering, executing a loop in physics simulation.}
    \label{fig:placeholder}
\end{figure}

\begin{abstract}
Robotic foundation models offer a promising path toward general-purpose humanoid robot control often with hierarchical architectures.
However, their effectiveness depends on a \textit{command interface} between the planner and the controller, which must support accurate execution while remaining easy to predict, and ideally compose new behaviors from prior ones.
In this work, we introduce a latent-representation of such interface we term \textit{spectral skills}, which meets these requirements through predictive representation learning. 
By design, the spectral skills compactly encodes short motion segments and are learned by predicting subsequent motion rather than reconstructing the encoder’s input. 
On a 29-DoF humanoid, a controller conditioned on spectral skills reduces global tracking error by $62\%$ relative to the state-of-the-art.
The same frozen controller chains independently encoded skills without a separate transition policy. It also composes new behaviors by adding orthogonal directions to any compatible base skill, producing combinations that are unseen from the training data.
We demonstrate tracking, chaining and composition, as well as through a languaged conditioned planner on Unitree G1 hardware.
Project page: \url{https://spectral-skill.github.io}.
\end{abstract}

\section{Introduction}

A central challenge in hierarchical humanoid control is deciding what should be communicated between a high-level behavior model and a low-level whole-body controller \citep{merel2018neural}. The high-level model must generate temporally coherent behaviors from abstract task specifications, while the controller must translate these commands into dynamically feasible actions at high frequency. 
Despite substantial advances in robot learning, modern systems largely preserve this hierarchical structure, replacing both components with learned models. 
At the high level, Behavior Foundation Models (BFMs), including Vision-Language-Action (VLA) models, learn to generate robot behaviors from large, diverse multimodal datasets. 
At the low level, reinforcement learning (RL) has emerged as a powerful tool for training robust whole-body controllers at scale \citep{luo2026sonic, wu2025learn}. 
Yet the command representation connecting these components remains underexplored. 
This representation must satisfy two requirements: it must be predictable by the high-level model and expressive enough to support diverse whole-body motions. Therefore we ask:
\emph{What command representation should a high-level planner use to instruct a whole-body controller?}

Classical hierarchical control methods implement this interface as explicit targets with physical meanings, leading to a wide range of interface designs. 
Some predict explicit full-body trajectories, joint-space targets, or velocity commands~\citep{kaelbling2011hierarchical, kuindersma2016optimization}, while others use lower-dimensional Cartesian targets such as pelvis and end-effector poses \citep{he2025hover, he2024omnih2o}. 
Similar hierarchical interfaces also appear beyond humanoid control, including robot manipulation \citep{belkhale2024rt}, dexterous manipulation \citep{chen2024object}, and navigation \citep{cheng2024navila}. 
Explicit interfaces have several appealing properties: they are interpretable, easy to supervise from motion data, and directly connected to conventional control objectives.
However, they can also expose the high-level model to the complexity of the robot's dynamics, especially those that predict robot joint targets, or end-effector Cartesian targets.
For example, for a high-degree-of-freedom humanoid, accurately predicting long-horizon joint or Cartesian trajectories can be a difficult learning problem, and the representation is inherently tied to a particular embodiment and command parameterization. Errors made by the high-level are accumulated and passed to the low-level controller, making robust and expressive closed-loop control challenging.

An increasingly popular alternative is to learn a \emph{latent interface} that compactly represents temporally extended motion sequences. 
Rather than predicting explicit targets in joint or Cartesian space, the high-level planner predicts a latent representation that conditions a whole-body controller.
Existing latent interfaces are typically learned through motion reconstruction or posterior inference \citep{liang2025clam, peng2022ase, luo2026sonic}.
Although these approaches demonstrate that low-dimensional latent spaces can support expressive whole-body behaviors, they leave three important questions open:
First, \emph{what should the latent interface capture}, for example the current pose of the segment or how the body will move next?
Second, \emph{how should it be learned} so that the controller can execute it accurately and the high-level model can predict it reliably?
Third, \emph{can the latent interface support composition}: can new behaviors be built from known skills, the way a person adds a wave to a walk, without data for every combination or retraining the controller? 

In this work, we study the latent interface from both sides of the hierarchy: how accurately a low-level controller can execute latent commands and how reliably a high-level behavior model can generate them in closed loop.
Drawing inspiration from predictive representation learning~\citep{gao2025spectral} and latent action models \citep{ye2025latent}, we introduce \textit{spectral skills}, a latent interface learned by factorizing the multi-step transition probability of reference motions. 
The resulting representation preserves information predictive of future motion outcomes, rather than merely reconstructing the motion segment from which it is encoded~\citep{luo2026sonic}. 
This design yields three key properties. 
\textbf{Expressivity:} the representation supports accurate and robust tracking across a diverse repertoire of whole-body motions. 
\textbf{Efficiency:} it can be learned independently of the controller using only offline data and provides a compact prediction target for a high-level behavior model. 
\textbf{Compositionality:} the spectral skill enters the predictor linearly, so we can read off \textit{spectral directions} that each change one and only one aspect of the motion, such as raising an arm or turning, and add them to a base skill.

In summary, our contributions in this work are as follows:
\begin{itemize}[leftmargin=*,itemsep=0pt,topsep=1pt,parsep=0pt]
\item \textbf{Spectral skills}: we train a latent skill space learned offline by predicting the motion that follows a segment.
\item \textbf{Expressive and efficient}: conditioned on spectral skills, one controller lowers global tracking error by $62\%$ relative to SOTA with $12\times$ fewer environment steps, and as a language planner's output, spectral skills raise success from $77.1\%$ to $91.1\%$ over predicting explicit trajectories .
\item \textbf{Composable}: the frozen controller composes new skills zero-shot by steering a base skill along spectral directions, producing combinations unseen from the data for the first time in whole body control. Extensive experiments in simulation and on a Unitree G1 in the real world demonstrate the effectiveness of our method.
\end{itemize}

\section{Related work}

\subsection{Humanoid Whole-Body Control}

Recent work in humanoid control has increasingly adopted motion-tracking reinforcement learning \citep{gu2026humanoid}. Rather than crafting task-specific rewards, these approaches learn to follow reference motions retargeted from motion-capture data, by optimizing rewards that encourage the robot to track the reference motion~ \citep{chen2025gmt,yin2026unitracker,wang2026omnixtreme,luo2026sonic}.
At inference time, however, they still require a specification of the desired motion, which is often difficult to specify directly. Existing approaches therefore commonly use a high-level planner to generate reference trajectories or latent representations from language and visual inputs for the low-level policy to track  \citep{jiang2025uniact,xie2026textop,tao2026heracles,zhang2026learning,peng2022ase,tessler2023calm,luo2024universal,
tessler2024maskedmimic,yao2024moconvq,luo2026sonic,
tirinzoni2025zero,li2026bfm}. The representations serve as commands to the low-level policy, allowing it to produce different behaviors without explicit reference trajectories. The high-level planners can then be supervised by predicting these commands \citep{xue2025leverb,jiang2026wholebodyvla,luo2026sonic,liao2026beyondmimic,zeng2026scaling}, which tends to be easier due to the semantic structure and low-frequency nature of these command interfaces. 
Like these methods, we learn a command space from motion data, but by predicting how motion continues rather than reconstructing it. 

\subsection{Predictive Latent Representations for Control}

The training procedure for our skill encoder is closely related to approaches that learn representations by factorizing system dynamics. 
Spectral representation methods \citep{gao2025spectral} use a low-rank factorization of the state–action transition kernel to obtain representations that are provably sufficient for downstream reinforcement learning \citep{ren2022free,zhang2022making,ren2023latent,ren2023spectral,zhang2023provable}. 
DiffSR \citep{shribak2024diffusion} learns such a factorization as an energy-based model through the score of a diffusion model, which captures richer transition distributions. 
Successor features \citep{dayan1993improving,kulkarni2016deep,barreto2017successor} instead factorize the cumulative state-visitation operator $(I-\gamma T^\pi)^{-1}$ under a behavior policy $\pi$, a multi-step operator, using its left singular vectors as features. 
Building on this idea, forward–backward representations jointly learn a low-rank factorization of future visitation probabilities and a family of latent-conditioned policies~\citep{touati2021learning}. 
Together, these works show how predictive representations can support reusable behaviors. 
We factorize the operator one level up the hierarchy: the transition of the high-level semi-MDP, that is, the law of the motion that follows when a skill runs for multiple steps learned with the diffusion recipe of DiffSR.

\subsection{Skill Composition}

Composition by adding energies has a long history: a product of experts multiplies densities, so their energies add~\citep{hinton2002training}, and energy-based and diffusion models compose concepts by summing energies or scores during sampling~\citep{du2020compositional,liu2022compositional}. 
Our model is an energy-based model in which the skill enters affinely, so composition happens in the skill space itself: adding spectral directions to a base skill adds their predicted effects. 
Extracting the directions of largest response from a Jacobian also has a long history.
Manipulability analysis uses the singular vectors of a manipulator's Jacobian~\citep{yoshikawa1985manipulability}, active-subspace methods use the leading eigenvectors of the expected Jacobian Gram of a vector-valued function~\citep{zahm2020gradient}, and in generative models semantic directions are computed in closed form from the first affine layer of a GAN~\citep{shen2021closed} or from the pullback metric of a generative model's decoder~\citep{shao2017riemannian} or of a diffusion model's features~\citep{park2023understanding}. 
In these generative models the map from latent space to output is nonlinear, so these directions are exact only for the first layer at best or hold only locally.
In our decoder, the response to a skill change does not depend on the skill (Prop.~\ref{prop:latent-response}). 
For humanoid robots motion composition, Meta Motivo and BFM-Zero condition one policy on a prompt vector for tracking, goal reaching and reward optimization~\citep{tirinzoni2025zero,li2026bfm}. 
However such methods are constrained by the existent of frames of the desired composed motion. Whereas spectral directions are extracted from the trained model and have generalization capability across the skill space.

\section{Learning spectral skills}
\label{sec:method}
\begin{figure}[t]
    \centering
    \includegraphics[width=\linewidth]{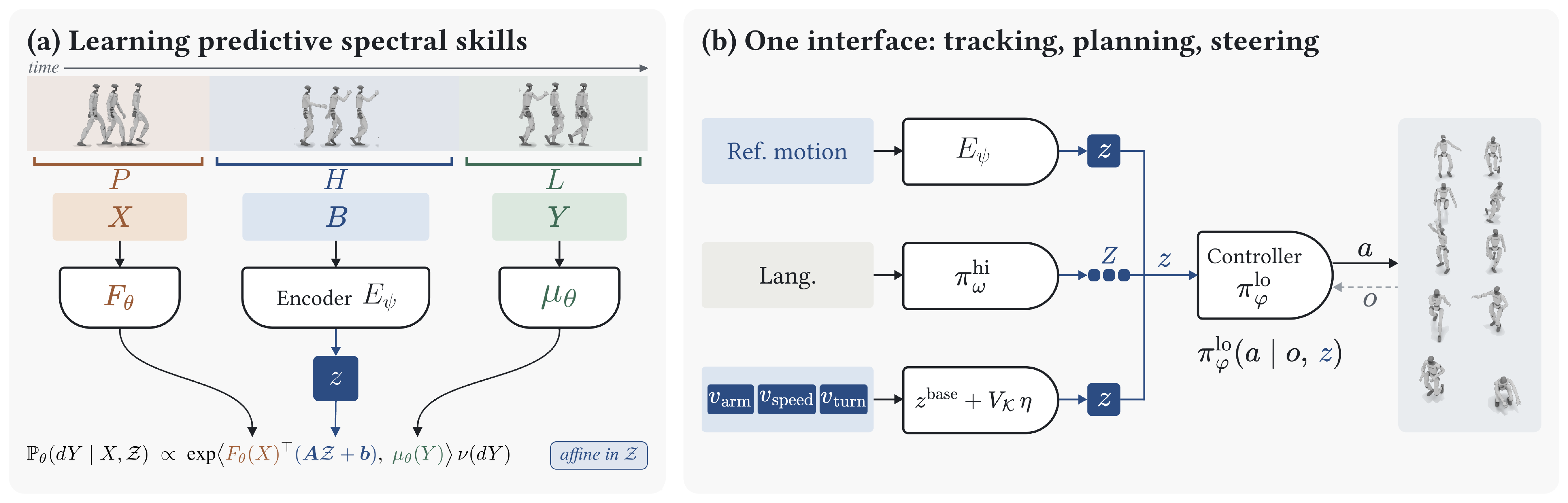}
    \caption{\textbf{Overview.} (a)~An encoder $E_\psi$ maps a state history
to a skill $z$ that linearly parameterizes a low-rank transition
model. (b)~One controller $\pi^{\mathrm{lo}}_\theta$ takes this skill from
reference tracking, language-conditioned planning, or skill composition.}
    \label{fig:banner}
    \vspace{-0.5cm}
\end{figure}

\paragraph{Preliminaries}
Figure~\ref{fig:banner} illustrates the overall framework. We formulate humanoid whole-body control as a hierarchical semi-Markov decision process (SMDP) operating at two time scales. The interface between them, which we call \textit{spectral skills}, is learned by factorizing transitions in offline motion capture data. Conditioned on a natural-language instruction, the high-level planner selects a skill to execute over an extended horizon; while a low-level policy trained through motion-tracking reinforcement learning converts the skill command into joint actions at a higher frequency.

Formally, consider the low-level MDP $\mathcal{M}^{\mathrm{lo}}=(\mathcal{S},\mathcal{O},\mathcal{A},\sP^{\mathrm{lo}},r^{\mathrm{lo}},\gamma)$. At micro step $t$, the state $s_t\in\mathbb{R}^{93}$ contains the joint positions and velocities, pelvis velocity, projected gravity, and previous action. The observation $o_t$ is derived from $s_t$ under domain randomization, and the action $a_t$ specifies a full joint command. Given a high-level command or spectral skill $z\in\mathcal{Z}$, the low-level policy selects an action according to $a_t\sim\pi_{\varphi}^{\mathrm{lo}}(\cdot\given o_t,z)$, parameterized by $\varphi$.

At macro step $k$, with decision time $t_k=k\Delta_t$, the high-level planner observes a macro state $X_k\in\mathcal{X}$, a window of $W$ recent robot configurations ending at $t_k$, where $x_t\in\mathbb{R}^{n}$ contains the reference motion. 
It selects a skill $z_k\in\mathcal{Z}$, which the low-level policy executes until the next decision, inducing the high-level transition $\sP^{\mathrm{hi}}(X_{k+1}\given X_k,z_k)$. The window length $W$ and the interval $\Delta_t$ are design choices and consecutive macro states may overlap or leave gaps.

\subsection{Learning Spectral Skills from Factorized Dynamics}

We learn the latent spectral skill with an encoder--decoder architecture. The encoder maps an executed motion segment to a compact skill representation, while the decoder combines this representation with the preceding context to predict the subsequent motion.

Let \(\mathcal{D}=\{(x_1^j,\ldots,x_{T^j}^j)\}_{j=1}^J\) be a dataset of \(J\) reference trajectories, where \(x_t^j\in\mathbb{R}^n\). For a sampled trajectory \(j\) and time \(t_k\), we divide the segment \(x_{t_k-P+1:t_k+H+L}^j\) into three consecutive windows:
\begin{equation}
X_k=x_{t_k-P+1:t_k}^j,
\qquad
B_k=x_{t_k:t_k+H-1}^j,
\qquad
Y_k=x_{t_k+H:t_k+H+L}^j.
\label{eq:motion-windows}
\end{equation}
Here, \(X_k\) is the observed context, \(B_k\) is the motion executed under a command, and \(Y_k\) is the subsequent motion. The parameters \(P\), \(H\), and \(L\) are the context length, command duration, and prediction horizon. The window \(Y_k\) contains \(L+1\) states, beginning at the boundary state \(x_{t_k+H}^j\), and does not overlap with \(B_k\). All windows are expressed relative to the frame at \(x_{t_k}^j\) before being passed to the networks.

\paragraph{Encoder.}
A deterministic encoder maps the executed motion \(B_k\) to a latent representation:
\begin{equation}
z_k=E_\psi(B_k)\in\skill.
\label{eq:encoder}
\end{equation}
We call \(z_k\) the \emph{spectral skill}: a low-d representation of the temporally extended motion in \(B_k\).

\paragraph{Decoder.}
\label{sec:decoder}
The decoder predicts \(Y_k\) from the context \(X_k\) and skill \(z_k\). Unlike reconstruction-based models such as variational autoencoders, it predicts the future rather than reconstructing \(B_k\). This encourages \(z_k\) to capture the aspects of \(B_k\) that determine its future dynamical consequences.

We construct the decoder by factorizing the high-level transition kernel as
\begin{equation}
\mathsf{P}(dY\given X,z)
\propto
\exp\left(
\left\langle
\phi(X,z),\mu(Y)
\right\rangle
\right)\nu(dY),
\qquad
\phi(X,z)=F_\beta(X)^\top(Az+b),
\label{eq:spectral-density}
\end{equation}
where \(\mu(Y)\in\mathbb{R}^r\) represents the outcome and \(\phi(X,z)\in\mathbb{R}^r\) represents the context--skill pair. Their inner product measures the compatibility of a future motion with the context and skill \citep{ren2022spectral}. The skill enters only through the affine map \(Az+b\), which is critical for learning a composable skill space, as discussed in Section~3.4.

Because the normalizing constant in \eqref{eq:spectral-density} is intractable, we follow \citet{shribak2024diffusion} and learn the decoder through diffusion-based score estimation. Given a noise schedule \(\{(\alpha_\tau,\sigma_\tau)\}_{\tau=1}^K\), we sample
\begin{equation}
Y^\tau
=
\alpha_\tau Y+\sigma_\tau\epsilon,
\qquad
\epsilon\sim\mathcal{N}(0,I).
\label{eq:noising}
\end{equation}
The score of \eqref{eq:spectral-density} factorizes as
\(\nabla_Y\log p(Y\given X,z)
=
(\partial\mu(Y)/\partial Y)^\top\phi(X,z)\).
Motivated by this structure, we parameterize the noise predictor as
\begin{equation}
D_\theta(X,z,Y^\tau,\tau)
=
M_\kappa(Y^\tau,\tau)^\top
F_\beta(X)^\top(Az+b),
\label{eq:affine-factorization}
\end{equation}
where \(F_\beta(X)\in\mathbb{R}^{e\times r}\), \(M_\kappa(Y^\tau,\tau)\in\mathbb{R}^{r\times m}\), \(A\in\mathbb{R}^{e\times d}\), \(b\in\mathbb{R}^e\), and \(m=n(L+1)\). The decoder parameters are \(\theta=\{\kappa,\beta,A,b\}\).

We jointly train the encoder and decoder using the standard noise-prediction objective:
\begin{equation}
\mathcal{L}_{\mathrm{pred}}(\psi,\theta)
=
\mathbb{E}\left[
\left\|
D_\theta(X,z,Y^\tau,\tau)-\epsilon
\right\|_2^2
\right],
\qquad
\mbox{where \quad} z=E_\psi(B).
\label{eq:diffusion-loss}
\end{equation}
The resulting \(z\) is a predictive representation of motion: the encoder extracts a spectral skill from \(B\), and the decoder uses it with \(X\) to predict \(Y\). Because \(z\) enters only through \(Az+b\), both \(D_\theta\) and the score estimate \(s_\theta=-D_\theta/\sigma_\tau\) are affine in \(z\) for fixed \((X,Y^\tau,\tau)\).

\subsection{Low-Level Motion Tracking and High-Level Planning}
\label{sec:tracking}

\paragraph{Low-level tracking.}
After achieving the skill encoder with offline data, we freeze the skill encoder and train a $50\,\mathrm{Hz}$ controller $\pi_\varphi^{\rm lo}$ with PPO to track reference motions conditioned on given skills. 
Let $o_t$ be the robot's proprioceptive observation. At each control step, the controller samples $a_t\sim\pi_\varphi^{\rm lo}(\cdot\given o_t,z_t)$, where $a_t\in\mathbb R^{29}$ specifies joint-position commands. 
During training, reference motions provide both the windows from which the frozen encoder extracts $z_t$ and the targets for the tracking reward. 
The training environment follows SONIC~\citep{luo2026sonic}; further details are given in Appendix~\ref{app:skill-protocol}.

\paragraph{High-level planning.}
The high-level planner maps the robot's last \(Q_{\rm hi}=10\) states and a language embedding to a sequence of \(N_{\rm hi}\) skill commands:
\begin{equation}
p_\eta^{\rm hi}(Z_t\given h_t,\ell),
\qquad
h_t=o_{t-Q_{\rm hi}+1:t},
\qquad
Z_t=(z_t,\ldots,z_{t+N_{\rm hi}-1}).
\label{eq:high-level-planner}
\end{equation}
We train the planner using the conditional flow-matching objective and GR00T action-head architecture~\citep{bjorck2025gr00t}. Each training target is a reference skill sequence \(Z_t^\star=(z_t^\star,\ldots,z_{t+N_{\rm hi}-1}^\star)\), where \(z_{t+j}^\star=E_\psi(B_{t+j}^{\rm obs})\) is computed by the frozen encoder from the reference states \(x_{t+j}\). Since \(h_t\) consists of \(o_t\), the planner is conditioned on the robot’s actual observation history rather than the reference trajectory \(x\), enabling closed-loop autonomous planning.

During inference, the planner samples \(\hat Z_t=(\hat z_t,\ldots,\hat z_{t+N_{\rm hi}-1})\sim p_\eta^{\rm hi}(\cdot\given h_t,\ell)\). The low-level controller executes the first \(N\leq N_{\rm hi}\) skills, one per control step, $a_{t+j}
\sim
\pi_\varphi^{\rm lo}
\left(\cdot\given o_{t+j},\hat z_{t+j}\right), 
0\leq j<N$, after which the planner is called again at \(t+N\) with the updated state history.

\subsection{Skill Composition through Spectral Steering}
\label{sec:skill-composition}

So far a spectral skill describes a motion we have seen: the controller tracks the skill of a walking clip and the robot walks. 
Humans, however, rarely reuse a motion as is. 
We wave while walking, turn while carrying a box, or lift our feet higher on uneven ground, and we do so without having practised every combination. 
Could a robot achieve the same? Could we take a base skill, change \emph{one} or \emph{several} aspect(s) of it, such as raising an arm, and leaving the rest of the motion intact, without new data and without retraining the controller?

Neither explicit joint targets nor a generic learned skill space supports this directly. 
Explicit joint targets must be changed in joint space, oftentimes outside the controller's model of the motion, so the change may compete with the controller instead of being carried out. 
In a generic learned skill space, the map from skills to motion is nonlinear and usually unknown, so neither the direction nor the size of a useful change is tractable.

Our model avoids both, because of the design choice in Section~\ref{sec:decoder}: the skill enters the predictor only through the affine map $Az+b$ (\eqref{eq:affine-factorization}).

\paragraph{Skill composition.}
Fix a context $\xi=(X,Y^\tau,\tau)$ with $\alpha_\tau>0$. From one denoising step, the model's estimate of the clean future motion is $\widehat Y_\xi(z)
=(Y^\tau-\sigma_\tau D_\theta(X,z,Y^\tau,\tau))/\alpha_\tau.$
Since $D_\theta$ is affine in $z$, so is this estimate, and the effect of any change of the skill has a closed form.

\begin{proposition}[Exact latent response]
\label{prop:latent-response}
For any fixed context $\xi$, the Jacobian of the denoised estimate with respect to the skill is
\begin{equation}
J_\xi
:=
\frac{\partial\widehat Y_\xi}{\partial z}
=-\frac{\sigma_\tau}{\alpha_\tau}
M_\kappa(Y^\tau,\tau)^\top F_\beta(X)^\top A.
\label{eq:latent-jacobian}
\end{equation}
It is independent of $z$. Consequently, for every $\delta z\in\mathbb R^d$,
$\widehat Y_\xi(z+\delta z)-\widehat Y_\xi(z)
=J_\xi\delta z$.
\end{proposition}
So in a given context, moving the skill by $\delta z$ moves the predicted future by exactly $J_\xi\delta z$, whatever the base skill. The identity concerns one denoising prediction at a fixed context. It does not claim that the full diffusion sampler, or the robot's executed motion, is globally linear; how closely the robot follows the prediction is measured in Section~\ref{sec:exp-composition}. The proof is in Appendix~\ref{app:skill-proofs}.

\paragraph{Spectral Directions.}
The Jacobian gives the effect of any change, but not which changes are useful. We look for directions in skill space that change the predicted motion strongly across many situations, not just in one. To do this, we sample $N$ contexts from recorded motions and average the squared response into the response Gram matrix
\begin{equation}
C=\frac{1}{N}\sum_{i=1}^{N}J_{\xi_i}^\top J_{\xi_i}.
\label{eq:response-gram}
\end{equation}
so that $v^\top Cv$ is the mean squared change in the predicted motion caused by a unit step $v$. The best such directions are its eigenvectors, which we call them Spectral Directions.

\begin{definition}[Spectral Directions]
\label{prop:spectral-directions}
Let $(\lambda_k,v_k)$ be the eigenpairs of $C$, ordered so that $\lambda_1\ge\cdots\ge\lambda_d\ge0$, with orthonormal eigenvectors. Then
$
v_k\in
\underset{\substack{\|v\|_2=1\
v\perp v_1,\ldots,v_{k-1}}}{\arg\max}
\frac{1}{N}\sum_{i=1}^{N}\|J_{\xi_i}v\|_2^2,$

and the maximum equals $\lambda_k$. Equivalently, the $v_k$ are the right singular vectors of the vertically stacked Jacobians, scaled by $1/\sqrt N$.
\end{definition}

Note that $v_k$ come from the trained model alone, without labels. 
The eigenvalue ranks a direction by how much it changes the predicted motion, not by how useful it is, so we examine directions across the whole spectrum and find what each one does by executing it with the frozen controller. 
Many turn out to change one recognizable thing: one arm rises, the feet lift higher, or the robot turns.

\paragraph{Composition is addition.}
With spectral directions in hand, composing a new skill is simple.
Given a base skill stream $z_t^{\rm base}$, for instance the codes of a walking clip, a set of directions $V_{\mathcal K}=[v_k]_{k\in\mathcal K}$, and time-varying amplitudes $\eta(t)\in \mathbb{R}^{\mathcal K \times 1}$, we steer the base skill as
\begin{equation}
z_t^{\rm steer}=z_t^{\rm base}+V_{\mathcal K}\eta(t).
\label{eq:latent-steering}
\end{equation}
Each component of $\eta(t)$ turns one direction on, off, up or down while the base skill continues, and by Proposition~\ref{prop:latent-response} the predicted effects of several directions add. 
Steering a walk with the arm direction and the turn direction, for example, asks for a turning walk with the arm raised, a combination the controller never saw as one clip. 
Similar motion exist, which is why the controller can still execute it, but the dataset does not contain motions that produce the steered skill exactly.

\section{Experiments}
\label{sec:experiments}

\begin{figure}[t]
    \centering
    \includegraphics[width=1.0\linewidth]{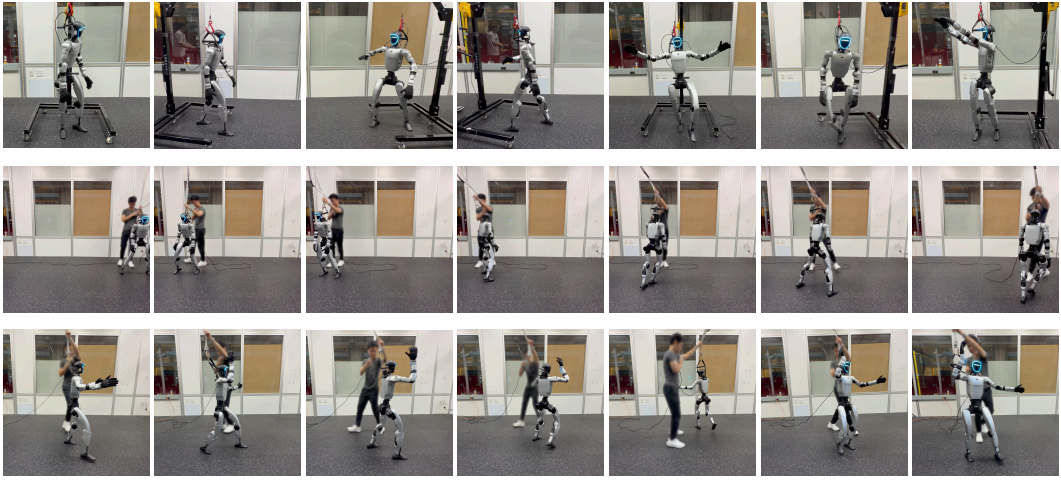}
    \caption{\textbf{Real-robot deployment of the frozen controller} on a Unitree G1. Top: tracking diverse whole-body motions. Middle: skill chaining, a side step, a $180^\circ$ turn and another side step, by switching skill codes. Bottom: knobs composed on one base skill. From left to right: base; right arm up; right arm up with increasing left turns (three panels) and increasing right turns (two panels) with each instance being a separate trajectory.}
    \label{fig:hardware}
    \vspace{-0.6cm}
\end{figure}

We evaluate whether spectral skills are expressive enough to drive whole-body motion in simulation and on hardware (Section~\ref{sec:exp-tracking}); whether they are composable, i.e., whether steering a base skill with spectral directions produces new skills (Section~\ref{sec:exp-composition}); and whether the hierarchy could benefit a language-conditioned planner to predict these skills (Section~\ref{sec:planner-interface}).
All experiments use a $29$-DoF Unitree G1, simulated in Isaac Lab~\citep{mittal2023isaaclab} with the Newton/MuJoCo-Warp backend~\citep{todorov2012mujoco} for tracking and planning and PhysX for rendering chaining and composition. 
Reference motions are the $129{,}785$ BONES-SEED clips~\citep{bones2026seed} released with SONIC~\citep{luo2026sonic}, with its G1 retargeting; the skill encoder is pretrained and frozen. Hardware, composition, and most chaining runs use a deployment version of the tracking controller, trained further with hardware-oriented regularization. We report success rate (SR) and mean per-joint position error in the root and world frames (MPJPE-L, MPJPE-G).

\subsection{Expressive Skills}
\label{sec:exp-tracking}

\begin{wraptable}{r}{6.6cm}
\vspace{-0.3in}
\caption{Whole-body tracking. $^1$Reported by SONIC on its own evaluation sets, not on these clips.}
\label{tab:tracking_main}
\centering
\footnotesize
\setlength{\tabcolsep}{3.5pt}
\begin{tabular}{@{}lrrr@{}}
\toprule
Method & SR $\uparrow$ & MPJPE-L $\downarrow$ & MPJPE-G $\downarrow$ \\
\midrule
Any2Track$^{1}$ & 69.4 & $\approx$60 & -- \\
BeyondMimic$^{1}$ & 85.4 & 39.1 & -- \\
SONIC$^{1}$ & 99.2 & 23.8 & -- \\
\midrule
\multicolumn{4}{@{}l}{\textit{124-motion capability set}} \\
SONIC & 100.00 & 23.79 & 173.92 \\
\textbf{Ours} & \textbf{100.00} & \textbf{18.22} & \textbf{65.06} \\
\midrule
\multicolumn{4}{@{}l}{\textit{4096-motion evaluation set}} \\
SONIC & \textbf{98.88} & 26.74 & 187.86 \\
\textbf{Ours} & 98.27 & \textbf{20.54} & \textbf{70.51} \\
\bottomrule
\end{tabular}
\vspace{-0.1in}
\end{wraptable}

\paragraph{Lower tracking error than SONIC.}
Conditioned on spectral skills, our controller tracks more accurately than the released SONIC tracker~\citep{luo2026sonic}, evaluated under identical settings on a $124$-motion capability set and $4{,}096$ motions across the corpus (Table~\ref{tab:tracking_main}; Appendix~\ref{app:eval-protocol}). 
The table also lists SONIC's reported numbers (its MPJPE is our MPJPE-L) for itself, Any2Track~\citep{zhang2025track} and BeyondMimic~\citep{liao2026beyondmimic}.
On both sets our controller lowers MPJPE-L by $23\%$ and MPJPE-G by $62$--$63\%$; on the 4096-motion set its SR is $0.6$ points lower, trading a little robustness on the hardest motions for trajectory fidelity. 
SONIC reproduces body configurations accurately but drifts from the commanded trajectory; root drift is $98\%$ of its squared global error (Appendix~\ref{app:sonic-tracking}). 
Our lower drift is consistent with a representation trained to predict how motion progresses over long horizons, and matched-budget ablations point to predictive supervision as the main source of the gain.
Our controller was trained on one GPU for $5.0\times10^{10}$ simulated frames, about $12\times$ fewer than SONIC.
Additionally, we conduct extensive ablation studies on the design choicesin Appendix~\ref{sec:ablation}.

\paragraph{Skill chaining.}
A controller that tracks skills can also chain them: switching the skill sequence from one motion to another mid-run moves the robot from one behavior to the next. On the real robot, the deployment controller tracks diverse motions and chains a side step, a $180^\circ$ turn and another side step (Figure~\ref{fig:hardware}, top and middle). 
In simulation, we compare the smoothness of such switches between our controller and SONIC, given the same motion sequences. We run $42$ chained sequences, in which each switch from one motion to the next takes between $2$ and $30$ control steps, and measure the root-mean-square joint acceleration during the switches (lower is smoother). Our switches are smoother in $36$ of the $42$ sequences, including all $13$ that switch within two steps. 

\subsection{Composable Skills}
\label{sec:exp-composition}
When steer a motion, the deployment controller and encoder stay frozen; a base clip is encoded into its base skill sequence $z_t^{\rm base}$, which we steer by adding spectral directions with steering amplitude $\eta$ (Equation~\plaineqref{eq:latent-steering}), and each steered run is compared with the unsteered run of the same base. 
The five bases (walk, jog, squat, carry, backward walk) differ in contact pattern (Appendix~\ref{app:composition-protocol}).

\begin{table*}[t]
\centering
\caption{Effects on the walk base at $a{=}2$ in Isaac PhysX.}
\label{tab:knob-catalogue}
\small
\setlength{\tabcolsep}{4pt}
\begin{tabular}{@{}ll|ll@{}}
\toprule
Direction & Measured effect & Direction & Measured effect \\
\midrule
Left arm raise  & shoulder pitch $\pm0.86$\,rad  & Arms out      & roll $+0.45$\,rad/side, $+0.43$ (carry) \\
Both arms       & shoulder pitch $\pm0.67$\,rad  & High steps    & swing apex $15\to36$\,cm (L), $14\to32$ (R) \\
Right arm raise & shoulder pitch $\pm0.90$\,rad  & Turn          & $\pm130^\circ$ in 4\,s \\
Right arm yaw   & shoulder yaw $\pm0.43$--$0.51$\,rad$^{\dagger}$ & Speed    & $+0.33$\,m/s, $6^\circ$ heading change ($a{=}3$) \\
\bottomrule
\end{tabular}
\end{table*}

\begin{figure}
    \centering
    \includegraphics[width=1\linewidth]{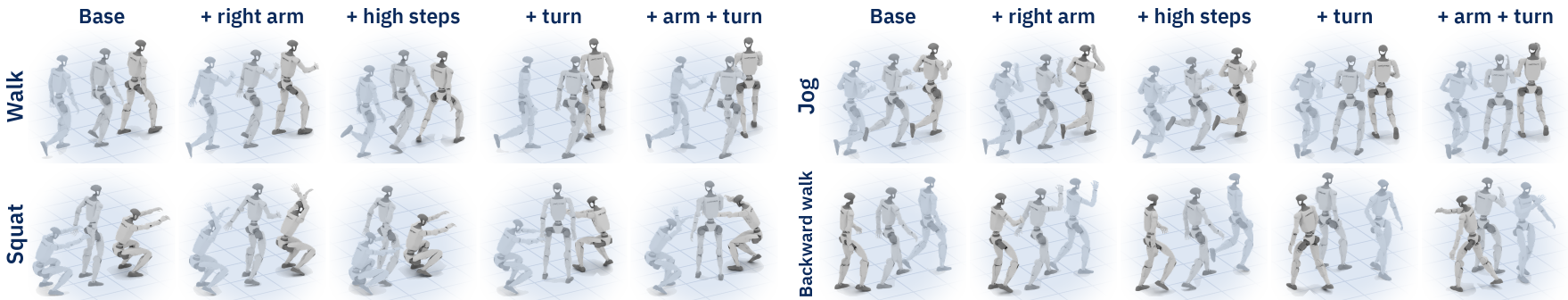}
    \caption{\textbf{One controller, many composed skills.} Rows: base skills (walk, jog, squat, backward walk); columns: knobs added to the same frozen controller.}
    \label{fig:composition-tiles}
    \vspace{-0.3cm}
\end{figure}

\paragraph{Steering.}
Adding a spectral direction to a base skill mainly changes one attribute while the base motion continues (Table~\ref{tab:knob-catalogue}, Figure~\ref{fig:composition-tiles}). 
The right-arm direction raises the arm while the walk continues; the turn direction turns the walk and barely moves the arms. 
Steering also works on the real robot (Figure~\ref{fig:hardware}, bottom). 
Effects add up: two directions held together change the joints like the sum of their single-direction changes. 
Whether a direction exhibits a clear, opposite effect in the two signs without falling depends on the base. Of $19$ sampled spectral directions, $13$ directions can be cleanly applied on the walk skill while $7$ on the squat skill, and five arm directions among them can be applied on all five bases (see Appendix~\ref{app:composition-extras} for more detail).
Additionally we provide negative results on the baseline methods from BFM-zero and SONIC in the Appendix~\ref{app:composition-baselines}.

\begin{figure}[t]
    \centering
    \includegraphics[width=1\linewidth]{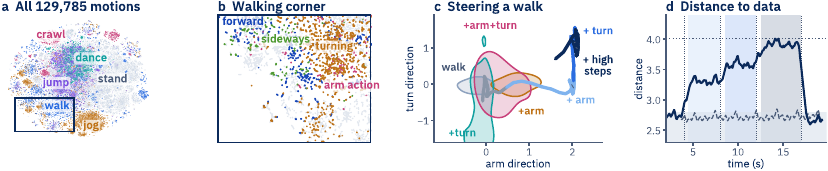}
    \vspace{-0.3in}
    \caption{\textbf{Steering composes skills rarely seen in the data.} (a) Map of all $129{,}785$ training clips (t-SNE), colored by motion kind; (b) walking corner. (c)--(d) One walk steered along the arm, turn and high-steps directions: (c) steered skill on the arm and turn directions, with contours of corpus walks that raise an arm or turn; (d) distance of the robot's motion to the corpus (dashed: unsteered) and corpus windows showing each combination. }
    \label{fig:tsne}
    \vspace{-0.1in}
\end{figure}

\paragraph{Latent space analysis.} 
Figure~\ref{fig:tsne} checks whether steering only retrieves motions that already exist in the training data. 
Panels (a) and (b) are maps of the training clips (corpus): each dot is one clip, placed so that clips with similar skills lie close together (t-SNE), and colored by the kind of motion in its name; 
(b) enlarges the corner where walks concentrate. 
Panels (c)--(d) follow one walk that we steer along the arm, turn and high-steps directions, added one at a time and then released. 
In (c), the axes are the skill's coordinates along the arm and turn directions: the steered skill moves right when the arm direction is added and up when the turn is added, and ends outside the regions where corpus walks that raise an arm or turn are most common (colored contours). 
In (d), the curve is the distance in the map from the robot's motion to the nearest corpus motion cluster over time: it rises with each added direction and falls back after the release, while the unsteered walk (dashed) stays at a typical distance towards the cluster center.

\subsection{Language-Conditioned Planning}
\label{sec:planner-interface}

\begin{wraptable}{r}{6.6cm}
\centering
\vspace{-0.5in}
\caption{Language-conditioned planning interfaces. Arrows denote planner output \(\to\) tracker command; \(z\) is skill and \(o\) is explicit root-and-joint frame. L/G: local/global MPJPE (mm).}
\label{tab:planner-interface}
\small
\setlength{\tabcolsep}{5pt}
\begin{tabular}{@{}lcrrr@{}}
\toprule
Interface & $N$ & L\,$\downarrow$ & G\,$\downarrow$ & SR\,$\uparrow$ \\
\midrule
\multicolumn{5}{@{}l}{\emph{Skill tracker: 256-D \(z\), hold 1}} \\
\quad Oracle                     & --- & 17.19 &  51.7 & 1.000 \\
\rowcolor{ctrlblue}
\quad \textbf{Skills (\(z\!\to\!z\))} 
                                 &  30 & \textbf{38.41} & 251.2 & \textbf{0.914} \\
\quad Skills (\(z\!\to\!z\))     &  10 & 42.83 & \textbf{230.2} & 0.911 \\
\quad Re-encode (\(o\!\to\!z\))  &  10 & 60.19 & 425.1 & 0.771 \\
\quad Re-encode (\(o\!\to\!z\))  &  21 & 88.64 & 796.8 & 0.546 \\
\midrule
\multicolumn{5}{@{}l}{\emph{Explicit tracker: 38-D \(o\), no encoder}} \\
\quad Oracle                     & --- & 17.64 & ---   & 1.000 \\
\quad Explicit (\(o\!\to\!o\))   &  10 & 116.46 & 704.9 & 0.525 \\
\quad Explicit (\(o\!\to\!o\))   &  30 & 117.42 & 707.1 & 0.516 \\
\bottomrule
\end{tabular}
\vspace{-0.3in}
\end{wraptable}

Skills can also serve as the action space of a high-level planner. We evaluate whether \(z\) provides a better interface between planning and tracking than explicit robot observations \(o\). In all routes, the planner predicts \(N_{\rm hi}=30\) commands, the controller executes the first \(N\), and the planner then replans from the updated robot history. Table~3 compares three routes, denoted by \emph{planner output \(\to\) controller input}, using the same planner architecture and training budget.

\emph{Skills (\(z\to z\)).}
The planner predicts a skill sequence
\(\hat Z_t=(\hat z_t,\ldots,\hat z_{t+N_{\rm hi}-1})\), and the continuous latent controller directly executes
\(a_{t+j}\sim\pi_\varphi^{\rm lo}(\cdot\given o_{t+j},\hat z_{t+j})\).

\emph{Explicit re-encode (\(o\to z\)).}
The planner instead predicts \(30\) explicit root-pose and joint-position frames
\(\hat O_t=(\hat o_t,\ldots,\hat o_{t+29})\). At step \(t+j\), the frozen encoder converts the predicted 10-frame window into
\(\hat z_{t+j}=E_\psi(\hat o_{t+j:t+j+9})\), which is then passed to the same continuous latent controller. Because each window must lie within the predicted chunk, this route supports only \(N\leq21\).

\emph{Explicit-to-explicit (\(o\to o\)).}
The planner again predicts \(\hat O_t\), but a controller trained without an encoder directly executes one predicted frame per step:
\(a_{t+j}\sim\pi_\varphi^{\rm exp}(\cdot\given o_{t+j},\hat o_{t+j})\).
Its planner is trained on rollouts from a separately trained explicit controller and is therefore not collection-matched with the other routes. For each controller, the \emph{oracle} replaces planner with the corresponding commands derived directly from the reference motion $x$, providing an upper bound on tracking.

At the matched replanning interval \(N=10\), Skills reduces MPJPE-L by \(29\%\) relative to Explicit re-encode and improves success from \(77.1\%\) to \(91.1\%\); Explicit-to-explicit achieves only \(52.5\%\) success. As \(N\) increases, MPJPE-L decreases for Skills but increases for Explicit re-encode. The ordering reverses at \(N=1\) (\(50.76\) vs.\ \(41.64\,\mathrm{mm}\) and \(84.3\%\) vs.\ \(94.5\%\) success), indicating that explicit predictions are effective under frequent replanning, whereas skills provide a more compact and robust interface over longer horizons. We also deploy the skill-based planner on hardware, where it produces smooth motion; videos are available on the project website.

\vspace{-0.3cm}
\section{Conclusions}
\vspace{-0.3cm}
We introduced spectral skills, a command space for humanoid whole-body control learned by predicting how motion continues, with the skill entering the predictor affinely. A controller conditioned on these skills tracks diverse motions with lower global error than SONIC and serves as the action space of a language-conditioned planner, and steering a base skill along spectral directions composes new skills that are rare in the data, without retraining.

\newpage

\subsection*{AI use statement}

In this work, we used generative AI tools to assist in the writing of the paper and assist in translation. 

We have not used generative AI tools to generate synthetic datasets, help develop theoretical models or conceptual frameworks, formulate mathematical claims, provide critical ingredients for proving mathematical claims, propose or refine hypotheses, design or provide feedback on research methodology or experiments, implement methods, clean and reformat datasets, support qualitative and thematic data analysis, or interpret results. 

Additionally, we used generative AI tools to organize and improve the visualization of the experiment results. 

We reviewed all AI-assisted content, including text and figures, to ensure that it accurately reflects the paper’s ideas and underlying data. We take full responsibility for the final content of this work, including any text, claims, or artifacts produced with the aid of generative AI.

\subsection*{Reproducibility statement}
We provide complete details of the model architecture, training objectives, reward design, environment configuration, and evaluation protocol in Appendix. Hyperparameters and additional ablations are also provided. We will release the training and evaluation code upon publication.

\bibliographystyle{iclr2027_conference}
\bibliography{iclr2027_conference}

\newpage

\clearpage
\appendix
\makeatletter
\@ifundefined{assumption}{\theoremstyle{definition}\newtheorem{assumption}{Assumption}\theoremstyle{plain}}{}
\providecommand\AppFloatBarrier{\par\begingroup\let\@elt\relax
  \edef\@tempa{\@botlist\@deferlist\@dbldeferlist}%
  \ifx\@tempa\@empty\else
    \ifx\@fltovf\relax\clearpage
    \else\newpage\let\@fltovf\relax\AppFloatBarrier\fi
  \fi\endgroup\suppressfloats[t]}
\makeatother

\paragraph{Appendix overview.}
\begin{itemize}
\item \ref{app:skill-geometry-control}: settings and proofs, body-part and target-directed directions, and the planner interface and its training.
\item \ref{app:expert-setting}, \ref{app:eval-protocol}: settings, metrics, motion sets, controllers and training compute.
\item \ref{sec:ablation}: interface ablations, and planning with an FSQ tracker.
\item \ref{app:sonic-tracking}: tracking against SONIC, whose world-frame gap is root drift.
\item \ref{app:composition-baselines}--\ref{app:composition-rigor}: composition against joint offsets, SONIC and BFM-Zero; more composition results and the experiment setting.
\item \ref{app:lafan1}: our method on LAFAN1 against BFM-Zero.
\end{itemize}

\section{Skill response: settings, proofs and planner}
\label{app:skill-geometry-control}

This appendix gives the settings and proofs behind spectral steering
(Section~\ref{sec:skill-composition}), the body-part and target-directed
directions, and the planner interface
(Appendix~\ref{app:planner-interface}). The exact skill response holds
only for the single-step denoising estimate, not for the sampler or the
robot rollout, so we qualify directions by execution.

\subsection{Windows, normalization and controller input}
\label{app:skill-protocol}

The analysis freezes one checkpoint; architectures and training budgets
are in Appendix~\ref{app:expert-setting}.

\paragraph{Windows.}
The encoder window $B_t$ of Equation~\plaineqref{eq:motion-windows}
shares $x_t$ with the context $X_t$. For the analyzed checkpoint
($H=L=10$), the boundary frame $x_{t+H}$ is the first target frame and
is hidden from the encoder.

Settings of the analyzed model:
\begin{center}
\begin{tabular}{ll}
\toprule
Quantity & Setting\\
\midrule
Reference state width $n$ & $29+3+6=38$\\
Past context $P$ & $6$ states\\
Encoder window $H$ & $10$ states, width $380$\\
Target horizon $L$ & $10$ steps; $11$ states, width $418$\\
Skill width $d$ & $64$\\
Predictor widths $(e,r)$ & $(1024,256)$\\
Regularization $(\beta,\lambda_z)$ & $(1,10^{-3})$\\
Reference sampling / controller rate & $50\,\mathrm{Hz}$\\
Skill refresh & One skill per control step\\
Controller history $Q$ & $10$ proprioceptive frames\\
Controller input width & $10\cdot93+64+2=996$\\
\bottomrule
\end{tabular}
\end{center}
Root positions and 6-D orientations are expressed in the heading frame
at the window origin. The $93$-D proprioceptive frame is listed under
\emph{Actor and critic} in Appendix~\ref{app:expert-setting}. The command appends a
sine/cosine phase pair to the $64$ skill coordinates. Steering
changes only the skill coordinates and keeps the phase.

\paragraph{Evaluation contexts.}
The analysis freezes all weights and normalization statistics and
samples past and future windows from recorded motions and queries the
lowest noise level $\tau_\star$ with zero added noise,
$y_{\tau_\star}=\alpha_{\tau_\star}y$. These fixed queries are not the
noisy distribution seen in training. Executing a spectral direction
needs no Jacobian or future window.

\subsection{Assumptions and proofs}
\label{app:skill-proofs}

Both propositions are exact for the frozen single-step estimate.

\begin{assumption}[Frozen affine response analysis]
\label{ass:affine-analysis}
The predictor has the form in Equation~\plaineqref{eq:affine-factorization}.
Its weights, target mean $m_Y$, and positive diagonal scale $S$ are fixed.
For each query $\xi=(X,y_\tau,\tau)$, differentiation holds all
components of $\xi$ fixed, with $\alpha_\tau>0$.
The set or distribution of queries does not depend on $z$.
The analyzed clean-target estimate is the unclipped single-step estimate
of Section~\ref{sec:skill-composition}, in the original motion coordinates:
\begin{equation}
  \widehat Y_\xi(z)=m_Y+\frac{S}{\alpha_\tau}\bigl(y_\tau-\sigma_\tau D_\theta(\xi,z)\bigr).
  \label{eq:app-single-step}
\end{equation}
\end{assumption}

\paragraph{Proof of Proposition~\ref{prop:latent-response}.}
At fixed $\xi$, collect the coefficient of $z$ in the noise field:
\begin{equation}
  D_\theta(\xi,z)=G_\xi z+c_\xi,\qquad
  G_\xi=\frac1{\sqrt e}M_\theta(y_\tau,\tau)^\top F_\theta(X)^\top A.
  \label{eq:noise-affine-coefficient}
\end{equation}
Substitution into Equation~\plaineqref{eq:app-single-step} yields
$\widehat Y_\xi(z)=d_\xi+J_\xi z$, with $z$-independent $d_\xi$ and
$J_\xi=-(\sigma_\tau/\alpha_\tau)SG_\xi$.
Proposition~\ref{prop:latent-response} states $J_\xi$ with $S$ and
$1/\sqrt e$ absorbed into $M_\theta$.
Subtracting its values at $z+\delta z$ and $z$ proves the identity for
every finite $\delta z$. There is no Taylor remainder. \hfill$\square$

The responses of different spectral directions are orthogonal on average
over contexts, not in each context, and for repeated eigenvalues only
the eigenspace is determined. The full sampler need not be affine in
$z$: in a reverse-diffusion update $y_{\tau-1}=T_\tau(y_\tau,z)$ the
noisy state $y_\tau$ itself moves with $z$, and clipping can change the
derivative further.

\subsection{Planner interface and sampling}
\label{app:planner-interface}

The planner reads $Q_{\rm hi}=10$ frames of $93$-D proprioception and
predicts $N_{\rm p}=30$ consecutive skills per call. For the continuous
tracker of Table~\ref{tab:planner-interface} ($256$-D, hold $1$), it
executes the leading $r_{\rm p}=N\in\{10,30\}$ skills at $50\,\mathrm{Hz}$
before replanning, i.e.\ at $5$ or about $1.7\,\mathrm{Hz}$.

The planner uses the GR00T N1.7 action head~\citep{bjorck2025gr00t}
with a frozen language backbone. Target skills are normalized with fixed
training-data statistics. The time $\tau$ in
Equation~\plaineqref{eq:planner-flow-matching} follows the head's
training schedule. An Euler sampler with
$K_{\rm hi}=4$ steps integrates
\begin{equation}
  \mathcal Z^{j+1}
  =
  \mathcal Z^j+\frac1{K_{\rm hi}}
  v_\omega(\mathcal Z^j,j/K_{\rm hi}\mid X_k,\ell),
  \qquad \mathcal Z^0\sim\mathcal N(0,I).
\end{equation}
Replanning times and histories are kept per environment.

\subsection{Planner training and evaluation}
\label{app:planner-training}

All planners in Table~\ref{tab:planner-interface} share one architecture
and training budget. The table compares routes at matched re-plan
intervals $N$, although each route has its own preferred interval. Only
the explicit-to-explicit rows use a different data collection.

\paragraph{Training.}
Each planner is the GR00T N1.7 action head of
Appendix~\ref{app:planner-interface}, trained for $12$k updates at batch
size $64$. The latent and explicit re-encode planners use the continuous
tracker of the table ($256$-D, hold $1$). We run that tracker on oracle
skills for $28$ language-described motions, with $93$ environments per
motion for up to $500$ control steps. Every step until the reference
ends gives one row, about $1.09$M rows per collection. A row pairs the
executed observation history with the skill chunk sent to the controller or, for the
explicit routes, with the $30$-frame root-and-joint target. The
explicit-to-explicit head was trained on the collection of an FSQ
tracker and drives an explicit tracker trained for $7.6\times10^{9}$
frames, so its rows are not collection-matched. 

\paragraph{Evaluation.}
We run $20$ episodes per motion ($560$ in total, seed $0$). Episodes
start on the reference, with observation noise but no pushes or domain
randomization, and end only on a fall or after $2{,}000$ steps. SR is the
fraction of episodes that never exceed SONIC's failure thresholds
(Appendix~\ref{app:eval-protocol}). MPJPE-L is averaged within each
episode and then over episodes.
For planner evaluation on hardware, we retrain the whole planner with a controller for a $64$-dimensional skill space. 
Everything else stays the same.

\AppFloatBarrier
\section{Experimental settings}
\label{app:expert-setting}

\emph{Ours} is the final tracker, trained for 50B environment frames. The
ablations of Appendix~\ref{sec:ablation} use their own training settings and are
scored at a matched 2B-frame checkpoint
(Table~\ref{tab:app-tracker-settings}), so compare each arm with the
ablation baseline, not with the 50B system.

\subsection{Network architectures}
\paragraph{Skill encoder.}
A reference frame holds 29 joint angles, the 3-D pelvis position and a
6-D pelvis orientation (38 values). Ten frames are flattened into a
380-D input. The encoder has four hidden layers of widths
$(2048,1024,512,512)$, each linear with LayerNorm and SiLU, and a linear
projection to $d=64$. The encoder is frozen during tracker training, except
in the posterior and online-adaptation variants.

\paragraph{Diffusion factorization networks.}
Ours uses affine conditioning (Equation~\plaineqref{eq:affine-factorization}) with feature width $r=256$ and embedding
width $e=1024$. The context $X$ is the current frame and five past frames
($6\times38=228$ values). The state network outputs
$F(X)\in\mathbb R^{1024\times256}$ and has residual hidden widths
$(512,512)$. The skill map is a single affine layer,
$g(z)=Az+b$. The noisy-target
network $M$ has residual hidden widths $(1024,1024,512)$ with Mish
and outputs $256\times m$ values for a target of width $m$. It sees the
noisy target and an embedding of the noise level $\tau$ (a 128-D
positional feature followed by $128\rightarrow256\rightarrow128$ linear
layers with Mish). The context enters only through $F(X)$.

A residual network with widths $(h_1,\ldots,h_J)$ maps its input to
$h_1$, applies $J$ blocks
\begin{equation}
 B(h)=W_2\operatorname{Mish}(W_1\operatorname{LN}(h)+b_1)
      +b_2+P(h),
\end{equation}
then Mish and a linear output layer. $P$ is the identity when widths
match and a learned linear map otherwise.

The original concatenation variant uses three residual networks, for the
state, the skill and their concatenated readout, each
with hidden widths $(1024,1024,512)$ and output widths 1024, 1024 and
256. Its context is a single 38-D frame.

\paragraph{Deterministic prediction and reconstruction heads.}
The next-skill predictor is
$d\rightarrow512\rightarrow512\rightarrow d$ with SiLU hidden layers.
The offline
reconstruction decoder mirrors the encoder,
$d\rightarrow512\rightarrow512\rightarrow1024\rightarrow2048
\rightarrow380$, with SiLU and a linear output.
The posterior route (the \emph{joint} rows of Table~\ref{tab:repr_ablation}) has its own encoder and reconstruction decoder, both
with two hidden layers of width 256 and ELU, and learns its 64-D skill
during tracker training, so it also differs from the offline routes in
architecture and data.

\paragraph{Actor and critic.}
Both have six hidden layers of widths
$(2048,2048,1024,1024,512,512)$ with SiLU. The actor outputs 29
joint-action means and a learned diagonal Gaussian standard deviation.
The critic outputs a scalar value from single-frame privileged inputs.
Ours stacks ten frames of projected gravity, base angular velocity,
joint positions, joint velocities and previous actions,
$10(3+3+29+29+29)=930$ values, and appends the 64-D skill and two phase
values (996 actor inputs). Observation normalization excludes the skill
input. The ablation baseline uses one proprioceptive frame
(159 actor inputs). Histories and command clocks reset independently in
each environment.

\subsection{Optimization and training budgets}
\begin{table}[tbp]
\centering
\small
\caption{Representation-learning settings.}
\label{tab:app-repr-settings}
\begin{tabular}{p{0.40\linewidth}p{0.54\linewidth}}
\toprule
Setting & Value \\
\midrule

Reference data & 129,785 retargeted BONES-SEED clips \\
Robot and simulator & 29-DoF G1; Isaac Lab with the Newton/MuJoCo-Warp backend \\
Control frequency & 50\,Hz \\
Offline pretraining & 50,000 updates; batch size 8,192 \\
Offline optimizer & AdamW; weight decay 0; gradient norm clip 1 \\
Encoder learning rate & $3\times10^{-4}$ \\
Generative next-chunk/next-skill head learning rate & $3\times10^{-4}$ \\
Deterministic next-skill predictor learning rate & $3\times10^{-4}$ \\
Separate endpoint DiffSR head learning rate & $10^{-4}$ \\
Offline reconstruction decoder learning rate & $3\times10^{-4}$ \\
Online posterior reconstruction learning rate & $3\times10^{-4}$ \\
Online dynamics adaptation learning rate & $3\times10^{-5}$ \\
Diffusion schedule & Variance-preserving; 8 noise levels \\
SIGReg & Coefficient 1; 64 random projections; 17 quadrature points \\
Bottleneck regularizer coefficient & $10^{-3}$, unless explicitly removed \\
Target-encoder EMA momentum & 0.996, where a next-skill target is used \\
\bottomrule
\end{tabular}
\end{table}

\begin{table}[tbp]
\centering
\small
\caption{Tracker-training settings.}
\label{tab:app-tracker-settings}
\begin{tabular}{p{0.40\linewidth}p{0.54\linewidth}}
\toprule
Setting & Value \\
\midrule

Tracker optimizer & AdamW; initial actor/critic LR $10^{-3}$ \\
Actor LR adaptation & Per-iteration KL; target 0.02; limits $[10^{-5},10^{-3}]$ \\
Ours: critic LR & Linear decay $10^{-3}\rightarrow10^{-5}$ over the 50B budget \\
Ours: tracker weight decay & $10^{-2}$ \\
PPO clipping; GAE; discount & $0.2$; $0.95$; $0.97$ \\
Entropy coefficient; gradient norm clip & 0; 1 \\
Ours: environments and rollout & 16,384 environments, 24 steps; batch 393,216 \\
Ours: PPO updates & 3 full-batch epochs per rollout \\
Ours: tracker budget & 50B cumulative environment frames \\
Ablations: environments and rollout & 20,480 environments, 24 steps; batch 491,520 \\
Ablations: PPO updates & 5 epochs; configured minibatch size 368,640 \\
Ablations: weight decay; critic LR & 0; constant $10^{-3}$ \\
Ablations: tracker budget & 5B frames per arm, seed 0; all rows scored at the matched 2.0B checkpoint \\
Command width; hold & 64 skill values plus 2 phase values; 1 control step \\
\bottomrule
\end{tabular}
\end{table}

The next-chunk diffusion head uses the encoder's
learning rate (Table~\ref{tab:app-repr-settings}). Only the separate endpoint head uses $10^{-4}$, and its
loss coefficient is zero in the final formulation. At runtime the skill
comes from the encoder alone. The pretraining heads are not run as
iterative denoisers at runtime.

Resets follow SONIC's adaptive sampler. Each clip is split into
$50$-frame bins. A clip and bin are drawn from a mixture that puts
weight $u$ on a uniform distribution and $1-u$ on the bins' normalized
failure rates. The episode starts up to $200$ frames before the drawn
bin. The 50B run lowers $u$ from 0.8 to 0.5 over the first 4B frames of
its first training stage and from 0.5 to 0.2 over the first 4B frames of
the second, then keeps 0.2. The share of failure-weighted resets thus
grows from $20\%$ to $80\%$. The ablation schedule ramps $u$ from 0.8 to
0.2 over its first 1B frames.

\subsection{Frames, windows and controller input}
\label{app:notation}
A \emph{frame} $x_t\in\mathbb R^{38}$ is one reference pose in a shared
anchor coordinate system. For anchor origin $o$ and rotation $R_a$,
\begin{equation}
 x_t^{(o,R_a)}=[q_t^*;R_a^\top(p_t^*-o);
                 \operatorname{rot6d}(R_a^\top R_t^*)].
 \label{eq:app-reference}
\end{equation}
The heading anchor uses a yaw-only rotation and the horizontal origin
$(p_x,p_y,0)$, so the reference height is kept. At runtime the anchor is
the robot's heading by default. Offline pretraining anchors on the
reference itself.

The skill $z$ of Section~\ref{sec:method} is the $d$-dimensional encoder
output for the window, or \emph{chunk}, $B_t=(x_t,\ldots,x_{t+H-1})$ of
Equation~\plaineqref{eq:motion-windows}, with horizon $H=10$ and stride $\delta=1$:
\begin{equation}
 z_t=E_\psi(B_t)=q\big(f_\psi(\operatorname{vec}(B_t))\big)\in\mathbb R^d,
 \label{eq:app-encoder}
\end{equation}
where $f_\psi$ is the encoder network and $q$ the bottleneck. The default skill is continuous
($q$ is the identity). The encoder sees $x_t$ through $x_{t+H-1}$ but not $x_{t+H}$.
For stride-$\delta$ variants, $x_{t+j}$ is the $j$-th sampled frame, at
$j\delta/50$\,s.

The \emph{next skill} (the target of the \textbf{Next-skill} arm in
Table~\ref{tab:predictive_target}) encodes the next non-overlapping chunk, which
starts at $x_{t+H}$ and is re-anchored to its own first frame. The
\emph{next-chunk diffusion target} of \emph{Ours} instead holds raw
future poses in the current chunk's anchor.

The \emph{command} is the held skill with its phase appended:
\begin{equation}
 c_t=[z_{t_{\mathrm{pub}}(t)};\sin(2\pi\rho_t);\cos(2\pi\rho_t)],\qquad
 \rho_t=(t-t_{\mathrm{pub}}(t))/K_{\mathrm{hold}},
 \label{eq:app-command}
\end{equation}
where $t_{\mathrm{pub}}(t)$ is the latest publication time in that environment and
$K_{\mathrm{hold}}$ is the hold period. At hold one the phase is the constant $(0,1)$.
Encoding horizon, prediction span and hold period are separate
quantities: a skill that summarizes ten frames can still be refreshed at
every control step.

\AppFloatBarrier

\section{Evaluation setup}
\label{app:eval-protocol}

Every BONES-SEED tracking result we measured, for ours and for SONIC, uses this setup.

\paragraph{Metrics.}
We report success rate (SR), local error (MPJPE-L) and global error (MPJPE-G).
As in SONIC~\citep{luo2026sonic}, an episode fails once the root or an end-effector height is off by more than $0.25$\,m, or the root orientation by more than $1$\,rad, and succeeds if it reaches its clip's end.
Let $p_{t,j}$ and $\hat p_{t,j}$ be the world positions of link $j$ at frame $t$ on the robot and the reference, over $14$ links $\mathcal{B}$: the pelvis ($j=0$), hips, knees, ankles, torso, shoulders, elbows and wrists.
Both errors are in millimeters, averaged over the frames $\mathcal{T}$ of successful episodes:
\begin{align*}
\text{MPJPE-L} &= \frac{1}{|\mathcal{B}||\mathcal{T}|}\sum_{t\in\mathcal{T}}\sum_{j\in\mathcal{B}}\big\|(p_{t,j}-p_{t,0})-(\hat p_{t,j}-\hat p_{t,0})\big\|, \\
\text{MPJPE-G} &= \frac{1}{|\mathcal{B}||\mathcal{T}|}\sum_{t\in\mathcal{T}}\sum_{j\in\mathcal{B}}\big\|p_{t,j}-\hat p_{t,j}\big\|.
\end{align*}
Only MPJPE-G penalizes drift off the reference path.
Both skip failed episodes, so compare them only at similar SR.

\paragraph{Motion sets.}
Both sets come from the $129{,}785$-clip BONES-SEED training corpus, with no clip held out.
The $4096$-motion set is sampled with a fixed seed after three filters that use only reference kinematics and clip names:
\begin{itemize}
\item[(i)] drop the $6{,}121$ clips that need absent objects or terrain (e.g.\ crates, doors, ladders);
\item[(ii)] drop clips shorter than $100$ or longer than $1{,}500$ frames, or with the pelvis below the floor;
\item[(iii)] drop the easiest quarter of the remainder by mean percentile of minimum pelvis height, peak root speed, 99th-percentile joint velocity and peak foot height.
\end{itemize}
Squatting, kneeling, crawling and boxing clips remain.
The $124$-motion capability set is selected differently (below).

\paragraph{Capability set.}
\label{app:capability-set}
The $124$-motion capability set of Table~\ref{tab:tracking_main} calibrates against SONIC and was chosen using SONIC's own results.
It starts from a separate SONIC evaluation on $4{,}096$ corpus clips, not the $4096$-motion set.
We removed clips with object or scene dependencies and clips outside $100$--$1{,}500$ frames, which left $3{,}732$ clips.
From these we drew seeded random $124$-clip subsets and kept the first one that meets coverage quotas and on which SONIC reaches an SR of at least $0.98$ and a success-only MPJPE-L of $23.5$--$23.9$\,mm, near its reported scale.
The quotas include at least $30$ locomotion, $15$ gesture, $5$ dance, $5$ jump or kick, $5$ injured-motion, $4$ ground-motion and $1$ boxing clip.

\paragraph{Training compute.}
\label{app:compute}
SONIC reports training its largest tracker for about $6.29\times10^{11}$ environment frames: $50$k iterations on $128$ GPUs, with $4{,}096$ environments per GPU and $24$ rollout steps per iteration, over approximately seven days (about $21$k GPU-hours)~\citep{luo2026sonic}.
Our tracker in Table~\ref{tab:tracking_main} was trained for $5.0\times10^{10}$ frames, about $12.6\times$ fewer ($16{,}384$ environments and $24$ steps per iteration; Table~\ref{tab:app-tracker-settings}).
It ran on one H200 GPU at a logged throughput of about $1.5\times10^{5}$ frames per second, in under $144$ GPU-hours.

\paragraph{Simulation settings.}
Tracking and planning run in Isaac Lab with the Newton/MuJoCo-Warp backend, and composition in Isaac PhysX.
Unless noted, runs use seed $0$, mean actions, no domain randomization or pushes, and the full clip from frame $0$.
Composition seed repeats add startup and reset randomization (Appendix~\ref{app:composition-rigor}), and the push test adds one push per $6$\,s run on top of that randomization (Appendix~\ref{app:composition-extras}).
Steering runs in the MuJoCo deployment simulator use the exported controller on a simulated G1 through the robot's deployment runtime at $50$\,Hz, with sensor noise measured on the real robot at rest.

\paragraph{Controllers.}
The paper uses three controllers; unless noted, each is conditioned on a $64$-D affine spectral skill.
\begin{itemize}
\item[(i)] \emph{Tracking} (Table~\ref{tab:tracking_main}, Appendix~\ref{app:sonic-tracking}): trained for $5.0\times10^{10}$ environment frames (Appendix~\ref{app:expert-setting}). It also runs four of the eleven chaining suites (Appendix~\ref{app:composition-protocol}).
\item[(ii)] \emph{Composition, hardware and the other seven chaining suites} (Section~\ref{sec:exp-composition}, Figure~\ref{fig:hardware}): the deployment controller, tracker (i) fine-tuned to $8.95\times10^{10}$ frames with a larger action-rate penalty, SONIC's power penalty, a larger joint-limit penalty, a joint-limit termination, a posture term, actuation delay, PD-gain and anchor randomization, and SONIC's reset-sampling values. The encoder and predictor are unchanged, so the same directions apply.
\item[(iii)] \emph{Planning} (Table~\ref{tab:planner-interface}): the continuous ($256$-D, hold $1$) and FSQ ($64$-D, hold $10$) trackers of an earlier interface study, each with its own planner. The hold is the number of control steps a skill is held.
\end{itemize}
Ablations (Appendix~\ref{sec:ablation}) are all scored at a matched $2\times10^{9}$-frame checkpoint.

\paragraph{SONIC.}
We run the publicly released SONIC tracker (v1.1, $41.6$M tracking parameters, the size of SONIC's largest model) in our environment, on our clips and metrics, re-encoding its token at every control step as in its deployment.
Its deployment set was never released, so the numbers SONIC reports for itself, Any2Track and BeyondMimic (marked in Table~\ref{tab:tracking_main}) come from its own evaluation sets, not from our clips.
Its measured rows in Table~\ref{tab:tracking_main} are scored by SONIC's own evaluator; Appendix~\ref{app:sonic-tracking} re-measures them with matched frame indices.

\AppFloatBarrier
\section{Ablation studies}
\label{sec:ablation}

We isolate the interface design choices behind the tracking results of Section~\ref{sec:exp-tracking}. Unless otherwise stated, every ablation arm is scored at the same checkpoint of $2\times10^{9}$ simulated environment frames (training continued to $5\times10^{9}$) and evaluated on the fixed 4096-motion set.

\paragraph{Prediction target.}
We first vary the prediction target while holding the tracker fixed. Our model predicts a complete future motion chunk. We compare against predicting only the horizon endpoint (\textbf{End-point}), a deterministic endpoint predictor (\textbf{End-point-det}), and predicting the skill of the next chunk (\textbf{Next-skill}).

\begin{table}[t]
\caption{
Prediction-target ablation. All arms, including \emph{Ours}, share one interface and are scored at the same $2.0$B-frame checkpoint on the same $4096$-motion set. \emph{Ours} here is not the 50B tracker of Table~\ref{tab:tracking_main}.
}
\label{tab:predictive_target}
\centering
\small
\setlength{\tabcolsep}{3.0pt}
\renewcommand{\arraystretch}{1.07}
\begin{tabular}{@{}lrrr@{}}
\toprule
Prediction target
& SR $\uparrow$
& MPJPE-L $\downarrow$
& MPJPE-G $\downarrow$ \\
\midrule

\rowcolor{ctrlblue}
\textbf{Ours}                  & 92.85 & 24.47 & 99.51 \\
End-point                      & 91.48 & 26.24 & 128.32 \\
End-point-det                  & 91.92 & 26.25 & 110.77 \\
Next-skill                     & 92.46 & 25.32 & 102.42 \\

\bottomrule
\end{tabular}
\end{table}

Predicting the complete future chunk gives the best overall result (Table~\ref{tab:predictive_target}): compared with endpoint prediction, MPJPE-G drops from $128.32$ to $99.51$\,mm and success rises from $91.48\%$ to $92.85\%$. A deterministic endpoint ($110.77$\,mm) and next-skill prediction ($102.42$\,mm) recover part of this gain.

\paragraph{Representation learning.}

We next vary how the skill space is learned. The \textbf{predictive} route (ours) pretrains the encoder with DiffSR and freezes it before RL.
The \textbf{reconstruction} baseline keeps the encoder architecture, input window, bottleneck and frozen-encoder pipeline, but trains the encoder offline as an auto-encoder that reconstructs its input window.
We also test whether a useful skill space can emerge from downstream control alone. In the \textbf{joint}
variants, no pretrained encoder is loaded; the encoder is optimized together with the tracking policy using reconstruction, policy gradient (PG), or both. Across these routes we also vary the
bottleneck: continuous, FSQ, categorical or VQ.

\begin{table}[t]
\caption{
Representation-learning ablation.
}
\label{tab:repr_ablation}
\centering
\small
\setlength{\tabcolsep}{2.5pt}
\renewcommand{\arraystretch}{1.06}
\begin{tabular}{@{}llrrr@{}}
\toprule
Representation & Training
& SR $\uparrow$
& MPJPE-L $\downarrow$
& MPJPE-G $\downarrow$ \\
\midrule

\rowcolor{ctrlblue}
\textbf{Ours}
& offline, cont. 64
& 92.85 & 24.47 & 99.51 \\

Ours
& offline, cont. 256
& 92.77 & 24.68 & 104.02 \\

Ours
& offline, FSQ
& 89.28 & 31.19 & 108.87 \\

Ours
& offline, cat.
& 68.48 & 46.97 & 273.29 \\

\midrule

Recon.
& offline, cont.
& 93.53 & 22.93 & 278.98 \\

Recon.
& offline, FSQ
& 93.68 & 22.50 & 238.12 \\

Recon.
& offline, VQ
& 5.10 & 62.79 & 909.34 \\

\midrule

Recon.
& joint, cont.
& 84.69 & 35.68 & 387.05 \\

Recon.+PG
& joint, cont.
& 88.92 & 30.82 & 418.26 \\

Recon.+PG
& joint, FSQ
& 86.67 & 30.69 & 420.71 \\

Recon.+PG
& joint, VQ
& 87.48 & 32.29 & 390.04 \\

\bottomrule
\end{tabular}
\end{table}

Both predictive and reconstruction encoders support good local tracking when pretrained before policy optimization (Table~\ref{tab:repr_ablation}). Reconstruction, however, yields much worse global tracking ($278.98$ vs.\ $99.51$\,mm MPJPE-G with continuous bottlenecks).

Training the encoder jointly with the policy is the weakest option for the viable continuous and FSQ bottlenecks. Moving the continuous reconstruction encoder from offline pretraining to joint training raises MPJPE-G from $278.98$ to $387.05$\,mm, and adding policy-gradient supervision raises it further ($418.26$\,mm). The joint FSQ variant reaches $420.71$\,mm, far above offline FSQ reconstruction ($238.12$\,mm) and predictive FSQ pretraining ($108.87$\,mm). Offline VQ reconstruction collapses ($5.10\%$ success), whereas joint VQ behaves like the other joint variants.

\paragraph{Interface design.}

Finally, we vary the remaining parts of the command interface with the predictive objective fixed. Table~\ref{tab:design_ablation} covers bottleneck geometry, encoder input and coordinate frame.

\begin{table}[t]
\caption{
Additional interface design choices.
}
\label{tab:design_ablation}
\centering
\small
\setlength{\tabcolsep}{2.7pt}
\renewcommand{\arraystretch}{1.05}
\begin{tabular}{@{}llrrr@{}}
\toprule
Design axis & Variant
& SR $\uparrow$ & MPJPE-L $\downarrow$ & MPJPE-G $\downarrow$ \\
\midrule

\rowcolor{ctrlblue}
\multicolumn{2}{l}{\textbf{Ours}}    & 92.85 & 24.47 & 99.51 \\

\midrule
\multirow{2}{*}{Skill width}
 & Cont. 128-D                       & 92.90    & 24.79    & 99.81     \\
 & Cont. 256-D                       & 92.77 & 24.68 & 104.02 \\

\midrule
FSQ & $64\times32$                   & 89.28 & 31.19 & 108.87 \\

\midrule
\multirow{3}{*}{Codebook}
 & Gumbel $64\times32$               & 69.60    & 44.03    & 180.60     \\
 & Cat. $64\times32$                 & 68.48 & 46.97 & 273.29 \\
 & VQ-EMA                            &  0.78 & 125.21 & 678.48 \\

\midrule
\multirow{5}{*}{Enc. input}
 & $+$ joint velocity                & 53.30 & 54.85 & 528.55 \\
 & Stride $5$                        & 73.56 & 37.17 & 147.36 \\
 & Full window                       & 92.63 & 24.97 & 111.85 \\
 & Horizon $5$                       & 91.58    & 25.93    & 96.84     \\
 & Horizon $20$                      & 56.69    & 54.93    & 635.31     \\

\midrule
\multirow{2}{*}{Anchoring}
 & Robot frame                       & 92.68    & 24.60    & 109.21     \\
 & Expert heading                    & 91.82 & 31.57 & 292.63 \\

\bottomrule
\end{tabular}
\end{table}

\textbf{Bottleneck.}
The continuous bottleneck is robust to its width: going from $64$ to $256$ dimensions changes MPJPE-G only from $99.51$ to $104.02$\,mm. FSQ remains viable but raises both local and global error. Learned discrete codebooks are much less stable: the categorical model drops to $68.48\%$ success and VQ-EMA nearly collapses.

\textbf{Encoder input.}
The encoder is sensitive to how time is presented. Sampling frames with stride $5$ hurts success and accuracy, and adding joint velocities is far worse than encoding a dense sequence of positions. Showing the full window keeps success but raises global error slightly. A $5$-frame horizon lowers MPJPE-G slightly ($96.84$\,mm) at some cost in success and local error, and a $20$-frame horizon degrades sharply.

\textbf{Coordinate frame.}
\emph{Ours} expresses the window in the live robot's heading frame (yaw only, horizontal origin). \emph{Robot frame} uses the robot's full anchor pose, which removes the reference's height and tilt relative to gravity. It costs little. \emph{Expert heading} anchors both pretraining and rollout on the reference's own heading, so the encoder never sees the robot's tracking error. Success stays high but MPJPE-G nearly triples.

\begin{figure*}[t]
    \centering
    \includegraphics[width=1.0\linewidth]{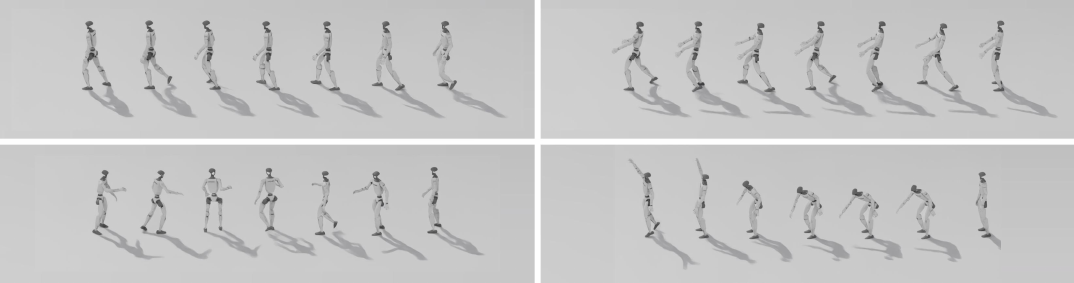}
    \caption{Language-conditioned rollouts. The planner produces skills that the low-level policy executes: walking in a circle, lifting and carrying a box, opening a door while turning, and reaching from low to high.}
    \label{fig:planner_qualitative}
\end{figure*}

\paragraph{Planning with an FSQ tracker.}
Table~\ref{tab:app-fsq-planner} repeats the planner comparison of Table~\ref{tab:planner-interface} with an FSQ tracker, which is not one of the 2B-frame ablation arms. The gap between latent and explicit prediction is smaller than with the continuous tracker. Temporal ensembling of latent predictions lowers both errors and raises success, so part of the planner error comes from inconsistency between successive predictions. Qualitative planner rollouts, not tied to this table, are shown in Figure~\ref{fig:planner_qualitative}.

\begin{table}[t]
\centering
\caption{Language-conditioned planning with an FSQ tracker ($64$-D, hold $10$). Every planner row re-plans every $N=10$ control steps (one held FSQ skill, or $10$ frames of the explicit chunk). Compare Table~\ref{tab:planner-interface}.}
\label{tab:app-fsq-planner}
\small
\setlength{\tabcolsep}{4pt}
\begin{tabular}{@{}lrrr@{}}
\toprule
Planner output & SR (\%) $\uparrow$ & MPJPE-L $\downarrow$ & MPJPE-G $\downarrow$ \\
\midrule
Oracle                               & --     & 20.47 & 90.5 \\
FSQ latent, temporal ensembling      & 90.4   & 45.49 & 220.5 \\
FSQ latent, no ensembling            & 87.7   & 52.30 & 246.5 \\
Explicit re-encode                   & 90.2   & 44.75 & 252.0 \\
\bottomrule
\end{tabular}
\end{table}

\AppFloatBarrier

\section{Comparison with SONIC: global root tracking}
\label{app:sonic-tracking}

SONIC tracks the body pose almost as well as ours, but its root drifts much further from the reference path.
This drift accounts for nearly all of the MPJPE-G gap in Table~\ref{tab:tracking_main}.
It grows faster with time than ours and lies mostly along the direction of travel.

\paragraph{Setup.}
We compare the tracker of Table~\ref{tab:tracking_main} with the released SONIC tracker~\citep{luo2026sonic}, run as described in Appendix~\ref{app:eval-protocol}.
We re-ran both motion sets, recording root positions and MPJPEs at every control step.
Scored by each method's own evaluator, as in Table~\ref{tab:tracking_main}, the re-runs match it to within $0.02$ SR points, $0.22$\,mm MPJPE-L and $0.5$\,mm MPJPE-G.
Below, both methods compare the state after control step $t$ with reference frame $t$.\footnote{SONIC's evaluator, used for its rows of Table~\ref{tab:tracking_main}, compares the state after step $t$ with reference frame $t-1$; ours already uses frame $t$, so our rows are unaffected. With matched indices, SONIC's MPJPE-L is $0.2$\,mm lower and its MPJPE-G $2.8$ and $2.9$\,mm higher on the $124$ and $4096$ sets.}

\begin{table}[t]
\centering
\caption{World-frame error decomposition, re-measured with matched frame indices (one seed, one evaluation per method). MPJPE and root drift $d$ are frame-weighted means over successful clips. Final is the median, over successful clips, of the root error at the last frame. All errors in mm. $^\ast$The $4{,}013$ clips of the $4096$ set that both methods complete.}
\label{tab:sonic-global}
\small
\setlength{\tabcolsep}{4pt}
\begin{tabular}{@{}llrrrrrr@{}}
\toprule
Set & Method & SR & MPJPE-L & MPJPE-G & Root drift $d$ & $\sqrt{d^2+\mathrm{L}^2}$ & Final \\
\midrule
124  & SONIC & 100.00 & 23.60 & 176.21 & 174.95 & 176.54 & 136 \\
     & Ours  & 100.00 & 18.00 &  65.05 &  58.66 &  61.35 &  15 \\
\midrule
4096 & SONIC &  98.88 & 26.52 & 190.82 & 189.27 & 191.12 & 177 \\
     & Ours  &  98.29 & 20.55 &  70.13 &  63.20 &  66.46 &  17 \\
\midrule
$4096^\ast$ & SONIC & -- & 26.35 & 187.7 & 186.1 & -- & -- \\
            & Ours  & -- & 20.48 &  70.0 &  63.1 & -- & -- \\
\bottomrule
\end{tabular}
\end{table}

\begin{figure}[t]
\centering
\includegraphics[width=\linewidth]{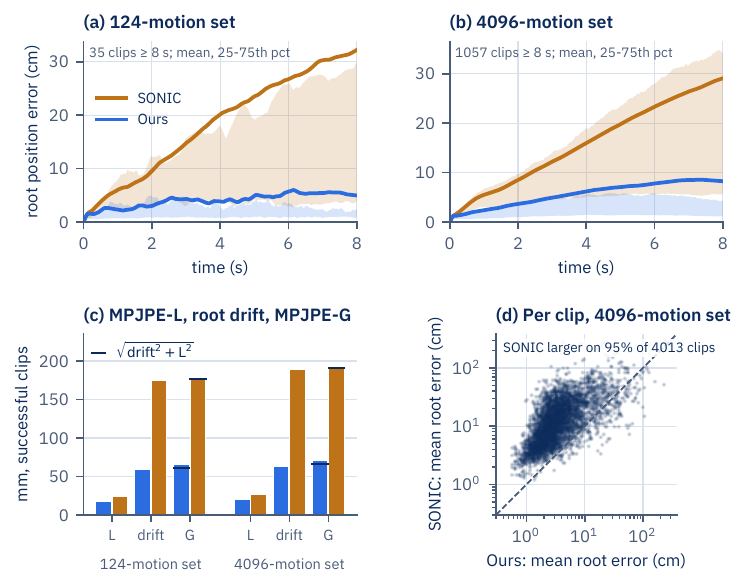}
\caption{Global root tracking, ours vs.\ SONIC. (a, b)~Root position error over the first $8$\,s of jointly completed clips lasting at least $8$\,s: mean (line) and interquartile range (band). In (a), a few SONIC clips that drift by over a meter lift its mean above the upper quartile. (c)~MPJPE-L, root drift and MPJPE-G over successful clips. The tick marks $\sqrt{\mathrm{drift}^2+\mathrm{L}^2}$. (d)~Mean root error per jointly completed clip of the $4096$ set. The dashed line is equality.}
\label{fig:sonic-global}
\end{figure}

\paragraph{Error decomposition.}
The world-frame error of each link (notation of Appendix~\ref{app:eval-protocol}) splits into a root term and a local residual:
\begin{equation*}
p_{t,j}-\hat p_{t,j} = \underbrace{(p_{t,0}-\hat p_{t,0})}_{\text{root drift}} + \underbrace{(p_{t,j}-p_{t,0})-(\hat p_{t,j}-\hat p_{t,0})}_{\text{local residual}} .
\end{equation*}
The root drift $d$ is the mean of $\|p_{t,0}-\hat p_{t,0}\|$ over the frames used for MPJPE.
If the two terms are uncorrelated, $\text{MPJPE-G}^2 \approx d^2 + \text{MPJPE-L}^2$.
This holds within $0.4$\,mm for SONIC and $3.7$\,mm for ours (Table~\ref{tab:sonic-global}).
Root drift makes up $98$--$99\%$ of SONIC's $\text{MPJPE-G}^2$ and $81\%$ of ours.
Between the methods, root drift differs by $116$--$126$\,mm and MPJPE-L by only $5.6$--$6.0$\,mm (Figure~\ref{fig:sonic-global}c).
The same holds on the clips both methods complete, which we call jointly completed ($4096^\ast$ rows).
Per clip, SONIC's mean root error is larger on $117$ of $124$ and $3{,}793$ of $4{,}013$ jointly completed clips (Figure~\ref{fig:sonic-global}d).

\paragraph{Drift over time.}
Figure~\ref{fig:sonic-global}a,b follows the root error over the first $8$\,s of the jointly completed clips lasting at least $8$\,s ($35$ clips of the $124$ set, $1{,}057$ of the $4096$ set).
A least-squares line fit to the mean root error has slope $36$\,mm/s for SONIC and $10$\,mm/s for ours on the $4096$ set ($40$ and $5$\,mm/s on the $124$ set).
At $8$\,s on the $4096$ set, the mean root error is $29.1$\,cm for SONIC and $8.3$\,cm for ours (medians $14.4$ and $2.0$\,cm).

\paragraph{Drift direction.}
On the $1{,}629$ jointly completed clips of the $4096$ set whose reference root ends at least $2$\,m from its start, we split the final horizontal root error (robot minus reference) into a signed part along the reference's start-to-end direction and an unsigned part across it.
SONIC ends short on $89\%$ of these clips, with median offsets of $28$\,cm along ($7.7\%$ of the net travel) and $11.5$\,cm across.
For ours the medians are $0.8$\,cm along and $1.2$\,cm across.

\begin{figure}[t]
\centering
\includegraphics[width=\linewidth]{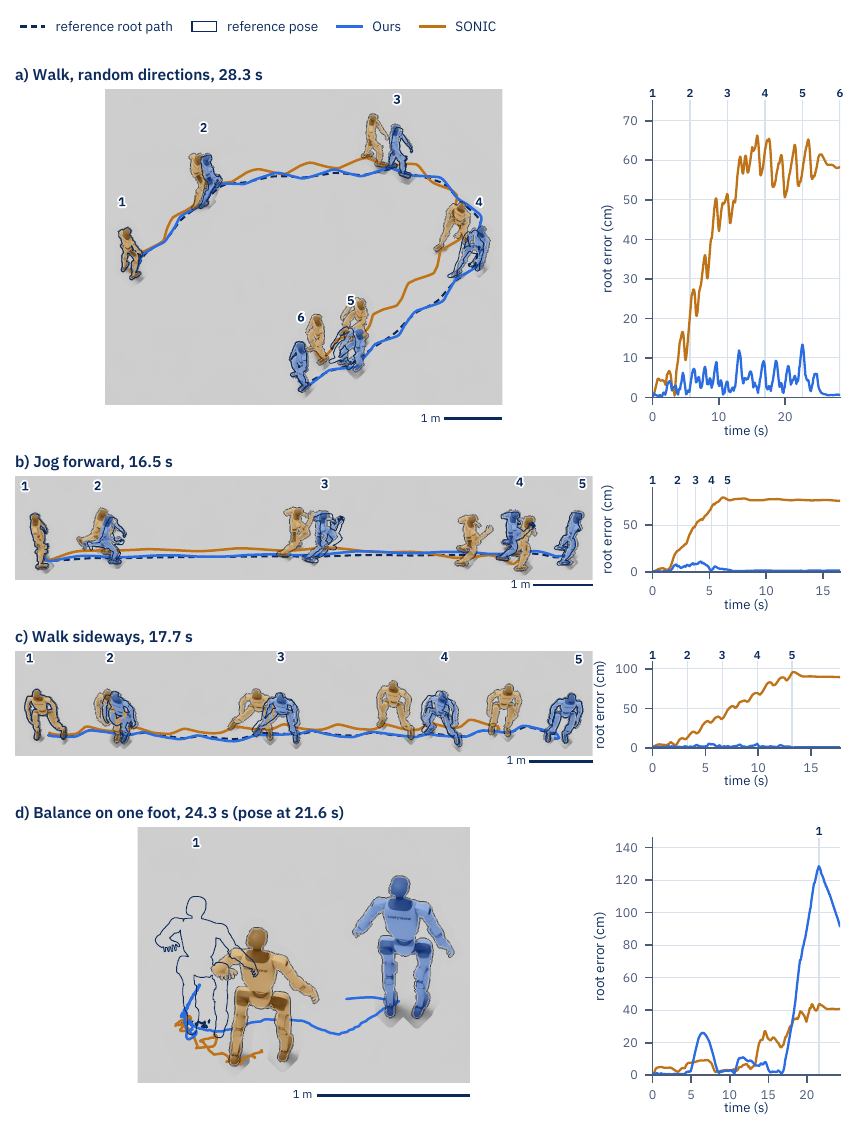}
\caption{Four long clips from the $4096$ set, seen from above. Left: rollouts, one fixed camera per clip. Root paths: reference dashed, ours blue, SONIC orange. Robots are drawn at the numbered times, ours and SONIC shaded, the reference as an outline. The scale bar is $1$\,m at the depth of the nearest path point. Right: root position error, with the pose times marked. In (b) and (c) the poses span the part of the clip in which the reference moves; (d) shows one pose, at the time of our largest error.}
\label{fig:sonic-birdview}
\end{figure}

\begin{table}
\centering
\caption{Root error of the four clips in Figure~\ref{fig:sonic-birdview}, averaged over the clip and at the last frame, in cm.}
\label{tab:sonic-clips}
\small
\setlength{\tabcolsep}{4pt}
\begin{tabular}{@{}llrrrr@{}}
\toprule
 & & \multicolumn{2}{c}{Mean root error} & \multicolumn{2}{c}{Final root error} \\
\cmidrule(lr){3-4}\cmidrule(l){5-6}
Clip & Reference motion & Ours & SONIC & Ours & SONIC \\
\midrule
(a) Walk, random directions & $28.3$\,s, $12.6$\,m path & 3.9 & 43.8 & 0.7 & 58.3 \\
(b) Jog forward & $8.9$\,m in $4.8$\,s, then stands & 2.7 & 62.1 & 1.6 & 76.2 \\
(c) Walk sideways & $8.3$\,m & 2.2 & 55.4 & 1.3 & 89.5 \\
(d) Balance on one foot & $24.3$\,s & 29.9 & 17.2 & 91.8 & 40.8 \\
\bottomrule
\end{tabular}
\end{table}

\paragraph{Qualitative examples.}
In clips (a)--(c) of Figure~\ref{fig:sonic-birdview} and Table~\ref{tab:sonic-clips}, SONIC's body matches the reference pose while its root is displaced.
In (b) and (c) the root lags mostly behind the reference along the path, and the offset persists after the reference stops.
For the renders, we re-ran each clip and replayed the recorded joint positions kinematically in Isaac PhysX, without alignment or re-timing.

\paragraph{Interface difference.}
We did not isolate the cause of the drift; one untested difference is that SONIC's released configuration gives the reference pelvis position in the robot's frame (\texttt{motion\_anchor\_pos\_b}) only to the critic, not to the actor or tokenizer.
Our encoder window includes the corresponding term (\texttt{expert\_anchor\_pos\_b}), so it reaches our actor through the skill.

\AppFloatBarrier

\section{Skill composition: comparison with alternatives}
\label{app:composition-baselines}

Steering changes both arm poses and gait. Joint offsets overshoot
or are canceled and cannot reproduce gait changes, SONIC~\citep{luo2026sonic}
token directions never reach a speed target, and BFM-Zero reward prompts
reach each target but with side-effect turns and a dependence on a
small part of its motion bank. Our runs follow the setting
of Section~\ref{sec:exp-composition} (Isaac PhysX, no domain
randomization, single seed).
Appendix~\ref{app:lafan1} compares ours and BFM-Zero on the same $40$ LAFAN1 clips, each with its own retarget and simulator.

\paragraph{Joint offsets.}
We replay each steered run's mean joint deviation as an offset under the
same policy, on the action (which the policy sees as its last action)
or on the servo target. The $66$ matched triples of
Table~\ref{tab:latent-vs-forcing} are drawn from the steered runs that
did not fall (five directions on five bases at three amplitudes, and one
wave gesture on walk).

\begin{table}[t]
\centering
\caption{Steering vs.\ matched joint offsets over $66$ triples
(5 bases, 5 directions, 3 amplitudes, and one wave gesture). Pose is the
achieved pose relative to the steered run, so it is $1.00$ for
steering by definition; steered runs have $0$ falls by construction. Speed is the median ratio to base
for arm directions on walk, jog and backward walk. Turn is the turn direction's
median heading change. The speed ratios are not robust across seeds
(Appendix~\ref{app:composition-rigor}).}
\label{tab:latent-vs-forcing}
\small
\setlength{\tabcolsep}{4pt}
\begin{tabular}{@{}lcccc@{}}
\toprule
Interface & Falls & Pose & Speed & Turn \\
\midrule
Steering                & 0/66  & 1.00 & 0.95 & $279^\circ$ \\
Offset on action        & 17/66 & 1.30 & 0.78 & $162^\circ$ \\
Offset on servo target  & 5/66  & 0.76 & 0.91 & $68^\circ$  \\
\bottomrule
\end{tabular}
\end{table}

Action offsets overshoot the pose because the policy feeds its last
action back in. The policy partly cancels servo offsets. Offsets
suffice for symmetric arm poses on slow or static bases (servo pose
ratio $1.03$ on squat) but not for gait. A copied foot-clearance
deviation crouches the robot (root $-7$\,cm) or topples it, and a copied
turn under-delivers. A copied waving gesture stalls
the walk ($0.01\times$ base speed on the action), while the steered wave
keeps $1.15\times$.

\paragraph{Pass rates.}
A run in Table~\ref{tab:knob-passrate} passes if it does not fall and
changes the target attribute by at least $g/2$, where the goal $g$ is
$1.0$\,rad for the right arm, $0.6$\,rad for spread (both arms out),
$+0.3$\,m/s for speed and $90^\circ$ for turn. Speed $v$ and heading $\psi$, unless targeted, must
stay near base: $|v/v_{\rm base}-1|\le0.3$ and
$|\psi-\psi_{\rm base}|\le27^\circ$.

\begin{table}[t]
\centering
\caption{Steering pass rates at $a\ge2$ on five bases (walk, jog, squat,
carry, backward walk). Target-directed directions are computed from chosen
output rows of $J_\xi$; spectral directions are
response-Gram eigenvectors. Joint offset (oracle) adds the matched steered
run's mean joint deviation to the servo target. Denominators count
runs: $10$ Isaac runs ($a{=}2$ and $3$ per base, one joint-offset spread
run lost), plus $2$, $2$ and $1$ walk runs of our right-arm, speed
and turn steering in the MuJoCo deployment simulator. Other rows have
one run per base at $a{=}2$ (SONIC in the MuJoCo simulator).
Spectral-direction and token-PCA rows use directions chosen on walk; single seed.}
\label{tab:knob-passrate}
\small
\setlength{\tabcolsep}{3.5pt}
\begin{tabular}{@{}lcccc@{}}
\toprule
Method & Right arm & Spread & Speed & Turn \\
\midrule
Ours, target-directed        & 9/12 & 9/10 & 4/12 & 6/11 \\
Ours, spectral directions only & 3/5 & 0/5  & 1/5  & 2/5  \\
Joint offset (oracle)        & 2/10 & 8/9  & 0/10 & 5/10 \\
SONIC, token PCA             & 4/5  & 2/5  & 0/5  & 0/5  \\
SONIC, ridge directions      & 0/5  & 0/5  & 0/5  & 4/5  \\
\bottomrule
\end{tabular}
\end{table}

Only our steering passes speed. With spectral directions alone, the
speed direction passes only on walk, the base it was chosen on. On the
arm targets, SONIC's label-free token PCA matches or exceeds our spectral directions, while its
label-fitted ridge directions fail on the arm and speed targets. Both representations
can turn, since both encode a heading-relative reference window.
Target-directed steering needs a predictor of future motion, which SONIC lacks.

\paragraph{BFM-Zero reward prompts.}
BFM-Zero~\citep{li2026bfm} has no joint or velocity command, so each
target is set by a reward prompt $r=r_{\rm walk}\cdot\tau$, where $\tau$ is a
tolerance on the target attribute. The released checkpoint infers its
prompt vector, which conditions the policy, by scoring this reward on the states of its LAFAN1 motion bank.
BFM-Zero runs in its own MuJoCo simulator on its own data, with a
deterministic actor and one seed.
Each run walks $4$\,s on the plain walk prompt, then $6$\,s on
the target prompt, of which we report the last $5$\,s (the hold window).
Literal speed and turn targets undershoot, so we calibrated them.
No run fell, and each prompt moves its target
(Table~\ref{tab:bfm-steering}). The arm raise also adds shoulder roll
and turns the robot.

\begin{table}[h]
\centering
\caption{BFM-Zero reward prompts applied to a walk (hold window, single seed). Yaw is the heading change over the window.}
\label{tab:bfm-steering}
\small
\begin{tabular}{@{}lrrrl@{}}
\toprule
Prompt & speed (m/s) & yaw ($^\circ$) & R.\ shoulder pitch (rad) & other \\
\midrule
Plain walk      & 0.60 & 22 & $+0.21$ & \\
Raise right arm & 0.69 & 81 & $-1.20$ & R.\ shoulder roll $-0.61$ \\
Spread arms     & 0.64 & 49 & $+0.29$ & roll L $+0.55$ / R $-0.50$ \\
Speed up        & 0.92 & 21 & $+0.27$ & \\
Turn            & 0.61 & 88 & $+0.22$ & \\
\bottomrule
\end{tabular}
\end{table}

\begin{figure}[t]
\centering
\includegraphics[width=\linewidth]{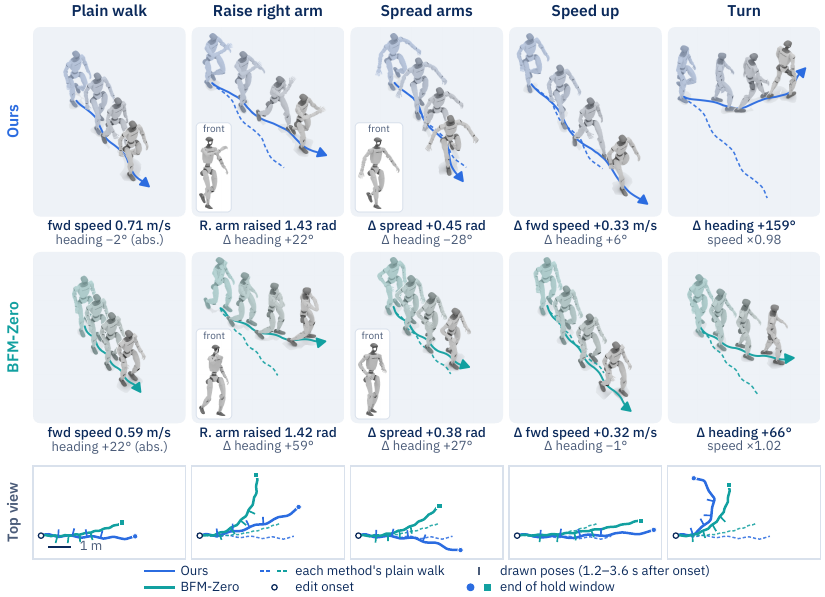}
\caption{A walk under our steering (top row) and under
BFM-Zero's reward prompts (middle row). Tiles show poses $1.2$--$3.6$\,s after
onset. The bottom row shows ground paths from above (left turns bend
upward). Numbers are changes against the same method's plain walk over
its hold window (ours $0.5$--$5.9$\,s, BFM-Zero $1.0$--$6.0$\,s after
onset); plain-walk tiles instead give that walk's own forward speed and
heading change. Spread is the mean outward shoulder-roll change of both
arms, and ``speed $\times$'' is the ratio of mean ground-path speeds. The
``front'' insets show the newest pose from the robot's front right.}
\label{fig:bfm-vs-ours}
\end{figure}

\paragraph{Matched changes.}
Figure~\ref{fig:bfm-vs-ours} runs the four targets with both methods on a
walk. For each target we use the steering amplitude whose change lies closest to
BFM-Zero's ($a{=}3$ for arm raise and speed-up, $a{=}2$ for spread,
$a{=}1.2$ for turn). The speed-up has the turn direction projected out.
The arm raise turns ours by $22^\circ$ and BFM-Zero by $59^\circ$
beyond its plain walk.

\paragraph{Arm swing.}
Over the hold window, the right shoulder pitch's standard deviation is
$0.03$\,rad for BFM-Zero and $0.27$\,rad for ours ($0.02$ and
$0.18$\,rad in the plain walks).

\paragraph{Side-effect turns.}
BFM-Zero's arm raise fails the pass rule on its turn, since its prompts carry no
heading term. Retrieval predicts
such side effects, but our arm raise also turns the walk, so these runs
cannot show how much of BFM-Zero's turn comes from retrieval.

\paragraph{Dependence on rare data.}
Removing the $0.85\%$ of bank frames that score a walk with both arms
up ($0.7$\,m/s, both wrists at least $1.0$\,m high, unlike the
right-arm prompt of Table~\ref{tab:bfm-steering}) stops BFM-Zero from
walking on that prompt ($0.76\to0.00$\,m/s). We rebuild our arm direction
from a response Gram over $100$ clips, after excluding every clip (and
its mirror) with a 1\,s window that walks, jogs, walks backward or
squats while a shoulder angle reaches $80\%$ of an arm direction's change
($5{,}508$ motions). The rebuilt direction's shoulder change goes from
$-1.43$ to $-1.34$\,rad on walk, $-0.85$ to $-0.81$ on jog and $-1.13$ to $-1.07$ on backward walk.
Our test removes clips before building the Gram, while BFM-Zero's removes frames at inference.

\AppFloatBarrier

\section{Additional composition results}
\label{app:composition-extras}

Runs use Isaac PhysX and one seed unless noted; the setup is in
Appendix~\ref{app:composition-protocol}.

\begin{figure}[h]
\centering
\includegraphics[width=\linewidth]{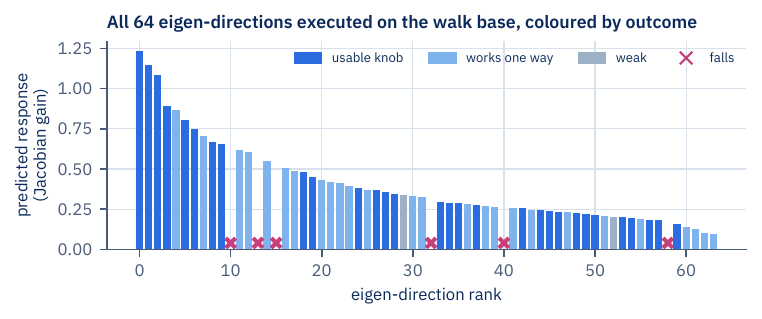}
\caption{All $64$ spectral directions (eigenvectors of the response
Gram) executed in both signs on the walk base. Bars give the predicted
gain by rank, colored by outcome. Crosses mark directions that fall in
both signs.}
\label{fig:all64}
\end{figure}

\paragraph{All 64 spectral directions.}
We executed every eigenvector of the response Gram in both signs on the
walk base, at an amplitude matched to its predicted gain
(Figure~\ref{fig:all64}). Of these, $30$ work in both signs, $26$ in
one, and the rest fall in both signs or are weak. Grouping directions whose
executed effects correlate at $|\rho|\ge0.7$ gives $22$ groups. The top
directions are single arm gestures. Ranks $11$--$64$ span a nearly
degenerate subspace: of these $54$ directions, $36$ have turning as
their dominant effect once amplified, two more drift sideways in one
sign and turn in the other, $7$ change speed, $6$ fall in both signs,
and $3$ move one limb or are weak.

\paragraph{Transfer across bases.}
Of $19$ representative directions, between $7$ (squat) and $14$
(carry), $13$ on the walk, work in both signs per base, including all five arm directions.

\paragraph{Steering vs.\ adding skills.}
Table~\ref{tab:skill-arithmetic} compares four ways to raise the right
arm during the walk; every line is applied for the whole run and no run
falls. A convex blend toward a standing clip with the right arm raised
trades walking for standing. Adding a skill difference keeps the walk
and raises the arm, but a motion that shows the raised arm also carries
the rest of its skill, so the difference brings other changes: the
difference from a standing idle turns the walk by up to $55^\circ$ and
lifts the left foot; the difference between the mean standing skills
with and without the raised arm also raises the left arm and both feet;
the same difference taken over walking skills turns and slows the walk.
Steering along the right-arm spectral direction moves the left arm by at
most $0.14$\,rad and keeps foot height within $3$\,cm of the unsteered
walk; its heading change grows with amplitude, to $41^\circ$ at $a{=}4$.

\begin{table}[h]
\centering
\caption{Raising the right arm during the walk (Isaac PhysX, walk base,
$400$ control steps, one seed; means over steps $65$--$335$). Blend:
$(1-\lambda)z^{\rm walk}+\lambda z^{\rm arm}$; skill differences:
$z^{\rm walk}+\lambda(z_B-z_A)$; steering: $z^{\rm walk}+a\,v$, with $v$
the right-arm spectral direction. Shoulder pitch is negative when the
arm rises.}
\label{tab:skill-arithmetic}
\small
\setlength{\tabcolsep}{4pt}
\begin{tabular}{@{}llrrrrr@{}}
\toprule
Line & $\lambda$ or $a$ & Speed & Heading & R.\ shoulder & L.\ shoulder & L.\ foot apex \\
 & & (m/s) & ($^\circ$) & (rad) & (rad) & (cm) \\
\midrule
Unsteered walk & -- & $0.71$ & $-2$ & $-0.05$ & $-0.04$ & $15.6$ \\
\midrule
Blend toward standing, arm raised & $0.5$ & $0.26$ & $-50$ & $-1.09$ & $+0.10$ & $8.0$ \\
 & $1$ & $0.00$ & $-11$ & $-2.38$ & $+0.27$ & $0.0$ \\
\midrule
Arm-raise clip $-$ standing idle & $0.5$ & $0.71$ & $-42$ & $-0.96$ & $+0.17$ & $20.6$ \\
 & $0.75$ & $0.66$ & $-55$ & $-1.64$ & $+0.33$ & $23.7$ \\
 & $1$ & $0.63$ & $-41$ & $-2.30$ & $+0.42$ & $25.6$ \\
\midrule
Standing: arm raised $-$ not raised & $0.5$ & $0.82$ & $-8$ & $-0.65$ & $-0.28$ & $18.4$ \\
(mean skills) & $1$ & $0.90$ & $+10$ & $-1.70$ & $-0.68$ & $22.9$ \\
 & $1.5$ & $0.94$ & $+15$ & $-2.65$ & $-1.03$ & $26.6$ \\
\midrule
Walking: arm raised $-$ not raised & $0.5$ & $0.75$ & $+27$ & $-0.59$ & $-0.35$ & $18.7$ \\
(mean skills) & $1$ & $0.44$ & $+77$ & $-1.56$ & $-0.91$ & $22.4$ \\
 & $1.5$ & $0.24$ & $+120$ & $-2.34$ & $-1.36$ & $25.9$ \\
\midrule
Steering, right-arm direction & $1$ & $0.74$ & $+4$ & $-0.46$ & $-0.08$ & $14.8$ \\
 & $2$ & $0.75$ & $+6$ & $-0.97$ & $-0.12$ & $14.6$ \\
 & $3$ & $0.73$ & $+20$ & $-1.51$ & $-0.15$ & $14.0$ \\
 & $4$ & $0.64$ & $+41$ & $-2.01$ & $-0.18$ & $12.7$ \\
\bottomrule
\end{tabular}
\end{table}

\begin{figure}[h]
\centering
\includegraphics[width=\linewidth]{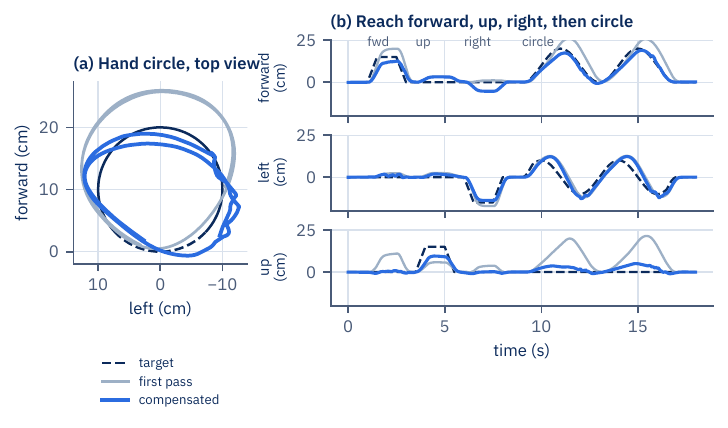}
\caption{Open-loop hand control on the standing base: (a) top view of a
traced circle, (b) hand position while reaching forward, up, right and
then tracing a circle. Dashed lines are targets, gray the first pass and
blue the compensated pass.}
\label{fig:ee}
\end{figure}

\paragraph{Hand targets.}
The hand Jacobian of the standing pose times the joint rows of the exact
skill response $J_\xi$ maps skill changes to hand motion. Its
regularized least-squares inverse, which also holds the legs, waist,
other arm and root, gives a $64\times3$ map that we apply open loop on
the standing base. The
first pass has a median error of $10.1$\,cm. One compensation pass,
which pre-inverts the measured linear map from target to achieved
hand displacement, lowers the error to $4.6$\,cm ($8.0$\,cm at the 90th percentile) without falls,
while the root drifts by up to $8.9$\,cm. A traced circle has $3.6$\,cm
error (Figure~\ref{fig:ee}).

\paragraph{Pushes.}
With startup and reset randomization and a push every $4$--$6$\,s (one
per $6$\,s run), the walk steered with the right-arm direction at
$a{=}1.8$ never fell in five seeds and kept $0.76$--$0.80$\,m/s, against
$0.69$--$0.73$\,m/s unsteered.

\subsection{Experiment setting and supporting measurements}
\label{app:composition-protocol}

\paragraph{Steering setup.}
Composition runs use the frozen encoder and the deployment controller
(controller (ii) in Appendix~\ref{app:eval-protocol}). Each base clip is
encoded open loop into its base skill sequence, and the steered sequence
is replayed to the controller. Replaying rather than re-encoding keeps
the base skill sequence fixed while steering changes the trajectory; a
base skill re-encoded in the robot's heading frame could undo a steered
turn. Steering ramps in and out over
$0.5$\,s ($25$ control steps), the same ramp as in the LAFAN1 runs of
Appendix~\ref{app:lafan1}. The steering amplitude $a$ is in raw skill
units. One unit is about $1.25$ standard deviations of the corpus skills,
whose mean per-dimension standard deviation is $0.80$. The five bases
are a forward walk, a jog, a squat, a two-handed carry that contains its
own $360^\circ$ turn, and a backward walk, each $n=390$--$400$ control
steps long. Steering runs from control step $40$ to $40$ steps
before the end, ramps included. Effects are averaged over the
full-amplitude part (steps $65$ to $n{-}65$) and compared with the
unsteered run of the same base. A run passes if the robot does not
fall, i.e.\ the root never drops below $0.40$\,m ($0.22$\,m on the
squat). The right-arm amplitude sweep and the high-steps entry of
Table~\ref{tab:knob-catalogue} come from an earlier $300$-step run of
the same walk, with the hold window at steps $65$--$235$.

\begin{figure}[h]
\centering
\includegraphics[width=\linewidth]{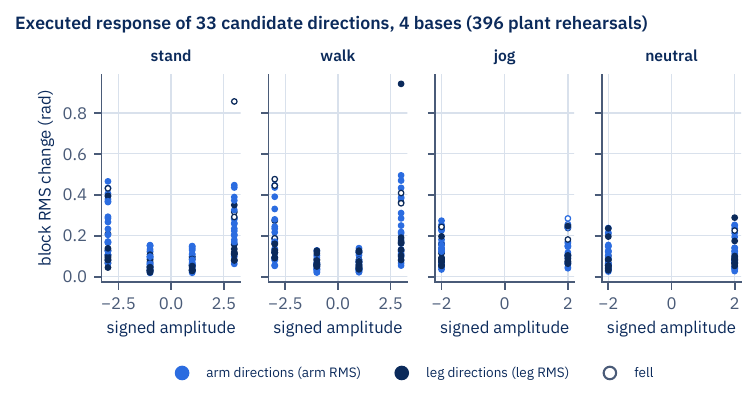}
\caption{RMS change of the targeted arm (blue) or leg (navy) joints
against signed amplitude. ``Stand'' is a standing typing idle and
``neutral'' a neutral standing idle. Open circles are falls.}
\label{fig:dose}
\end{figure}

\paragraph{Steering measurements.}
On the walk in Isaac PhysX, the right-arm direction reaches a shoulder
pitch of $-0.53$, $-0.83$ and $-1.45$\,rad at $a{=}1.2$, $1.8$ and
$3.0$. The walking speed is $0.77$, $0.79$ and $0.81$\,m/s, against
$0.72$\,m/s unsteered. In the MuJoCo deployment simulator, the turn
direction at $a{=}\pm2$ on a walk changes the heading by $+131^\circ$
and $-129^\circ$ over $4$\,s. The arms deviate by only $0.05$ and
$0.06$\,rad RMS, and the stride frequency stays at $0.79$\,Hz. Steering
effects add up in the same simulator. The right arm with a left turn
turns $+144^\circ$, against $+143^\circ$ for the sum of the two
single-direction runs (joint-pattern correlation $1.00$). Adding a lowered
stance gives $+152^\circ$ against $+141^\circ$ (correlation $0.98$). The
superposition cosines of Table~\ref{tab:latent-linearity} cover three
pairs, the right arm with a left turn, with a right turn and with a
lowered stance ($0.992$--$0.996$), and this three-direction case
($0.976$). Additivity degrades when directions overlap: with four
directions (left arm, right arm, turn, speed) scheduled with partial
overlaps, so that at most three are active at once, the turn delivers
$+15^\circ$ instead of $+85^\circ$ in a single full-amplitude run in
the deployment simulator.

\paragraph{Support and distance to the data.}
Each one-second corpus window ($1{,}697{,}723$ in total) is described by
its forward speed, yaw rate, root height and four shoulder angles.
Swing-foot apex is added for the sequence of Figure~\ref{fig:tsne}. The
support of an executed composition is the number of windows that move
like the base and in which every steered attribute has changed at least
$80\%$ as much as in the executed run. The eleven two-direction runs
pair an arm direction with the turn on the backward walk, jog, walk and
squat, and none falls. Their supports are $1$, $3$, $3$, $3$, $37$,
$38$, $53$, $96$, $117$, $132$ and $289$ windows. The lowest support of
a single direction is $2$ windows (backward walk with arms out). None of
these runs has zero support, whereas the backward-walk ladder of
Table~\ref{tab:ladder-seeds} reaches zero at its turn and clearance
stages. The steered skills of the single-direction five-base
runs have a median nearest-corpus distance of $3.42$, against $2.79$
between corpus skills (leave-one-clip-out, $99$th percentile $4.90$),
whereas Table~\ref{tab:ladder-seeds} measures the distance of the
re-encoded executed motion.

\AppFloatBarrier

\label{app:composition-rigor}

We repeat the composition ladder of Figure~\ref{fig:tsne} and the
$a{=}2$ runs of Table~\ref{tab:latent-vs-forcing} over three randomized
seeds. Both reproduce qualitatively, but at $a{=}2$ the speed cost of
action offsets does not hold.

\paragraph{Setup.}
All runs use the controller of Section~\ref{sec:exp-composition} in
Isaac PhysX. The deterministic run has no randomization, as in the main
text. Seeds $0$--$2$ add startup
randomization (friction $0.3$--$1.6$, per-joint offsets of the default
joint positions of up to $\pm0.01$\,rad, torso center of mass, wrist and
torso mass $\times0.8$--$2.5$) and a random start pose and velocity,
without pushes. Skills and the deterministic run's offset tables are fixed across
seeds, and the SD is the sample SD over the three seeds.

\begin{table*}[t]
\centering
\caption{Composition ladder over three randomized seeds (mean $\pm$
SD; brackets: deterministic run). Steering along the right-arm, turn
and high-steps directions starts at $4$, $8$ and $12$\,s ($a{=}2$, $1.2$
and $1.2$), and all three are released together at $17$\,s. Each row averages the steady control
steps of one stage. Speed is horizontal path speed, yaw the mean yaw
rate, shoulder the mean right shoulder pitch, and apex the swing-foot
height range over 1\,s windows (left/right). Distance is the median over
the stage of the Euclidean distance from the re-encoded executed skill to
the nearest of $263{,}395$ skills from $4{,}000$ corpus clips (corpus
skills to the nearest skill of another clip: median $2.80$, p90 $4.01$). Support counts corpus 1\,s windows (of
$1{,}697{,}723$) that move like the base and show every attribute moved
so far, one entry per seed. No run fell.}
\label{tab:ladder-seeds}
\scriptsize
\setlength{\tabcolsep}{3pt}
\begin{tabular}{@{}lcccccl@{}}
\toprule
Stage & Speed (m/s) & Yaw ($^\circ$/s) & Shoulder (rad) & Apex L/R (cm) & Distance & Support \\
\midrule
\multicolumn{7}{@{}l}{\emph{Forward walk}} \\
Base & $0.98{\pm}0.03$ & $-8{\pm}2$ & $0.05{\pm}0.01$ & $15{\pm}0$/$16{\pm}0$ & $2.73{\pm}0.09$ [2.76] & 90009, 92259, 93794 [90252] \\
+ right arm & $1.07{\pm}0.01$ & $3{\pm}1$ & $-0.76{\pm}0.02$ & $15{\pm}0$/$16{\pm}0$ & $3.32{\pm}0.04$ [3.37] & 2334, 1988, 2124 [2180] \\
+ turn & $1.06{\pm}0.03$ & $35{\pm}2$ & $-0.77{\pm}0.03$ & $14{\pm}1$/$15{\pm}0$ & $3.63{\pm}0.08$ [3.60] & 112, 102, 100 [119] \\
+ clearance & $1.11{\pm}0.06$ & $36{\pm}3$ & $-0.80{\pm}0.04$ & $19{\pm}2$/$24{\pm}1$ & $3.93{\pm}0.03$ [3.96] & 19, 35, 95 [21] \\
Released & $1.02{\pm}0.01$ & $-1{\pm}2$ & $0.01{\pm}0.02$ & $16{\pm}0$/$15{\pm}1$ & $2.87{\pm}0.15$ [2.69] & -- \\
\addlinespace
\multicolumn{7}{@{}l}{\emph{Jog}} \\
Base & $1.21{\pm}0.03$ & $2{\pm}1$ & $-0.10{\pm}0.01$ & $17{\pm}0$/$16{\pm}0$ & $2.95{\pm}0.03$ [2.93] & 68531, 68818, 71607 [71050] \\
+ right arm & $1.45{\pm}0.04$ & $3{\pm}1$ & $-1.10{\pm}0.04$ & $17{\pm}0$/$19{\pm}0$ & $3.66{\pm}0.03$ [3.65] & 245, 210, 276 [228] \\
+ turn & $1.32{\pm}0.03$ & $42{\pm}3$ & $-1.09{\pm}0.04$ & $16{\pm}1$/$17{\pm}1$ & $4.26{\pm}0.05$ [4.22] & 23, 13, 20 [21] \\
+ clearance & $1.49{\pm}0.14$ & $38{\pm}5$ & $-1.17{\pm}0.07$ & $27{\pm}1$/$32{\pm}1$ & $4.39{\pm}0.11$ [4.32] & 1, 0, 0 [1] \\
Released & $1.35{\pm}0.11$ & $-4{\pm}3$ & $-0.13{\pm}0.03$ & $18{\pm}1$/$17{\pm}1$ & $3.32{\pm}0.25$ [3.09] & -- \\
\addlinespace
\multicolumn{7}{@{}l}{\emph{Backward walk}} \\
Base & $0.84{\pm}0.00$ & $1{\pm}1$ & $-0.15{\pm}0.00$ & $11{\pm}0$/$10{\pm}0$ & $2.67{\pm}0.02$ [2.66] & 27865, 27582, 27902 [27626] \\
+ right arm & $0.76{\pm}0.02$ & $6{\pm}1$ & $-1.38{\pm}0.02$ & $11{\pm}1$/$13{\pm}0$ & $3.69{\pm}0.03$ [3.67] & 206, 193, 189 [192] \\
+ turn & $0.58{\pm}0.01$ & $45{\pm}2$ & $-1.36{\pm}0.01$ & $11{\pm}1$/$11{\pm}0$ & $4.13{\pm}0.03$ [4.10] & 0, 0, 0 [0] \\
+ clearance & $0.38{\pm}0.04$ & $51{\pm}7$ & $-1.47{\pm}0.06$ & $19{\pm}2$/$24{\pm}0$ & $4.17{\pm}0.02$ [4.15] & 0, 0, 0 [0] \\
Released & $0.73{\pm}0.05$ & $-5{\pm}3$ & $-0.29{\pm}0.00$ & $12{\pm}0$/$10{\pm}1$ & $2.70{\pm}0.11$ [2.65] & -- \\
\bottomrule
\end{tabular}
\end{table*}

\paragraph{Ladder.}
On every base, each direction moves its own attribute
(Table~\ref{tab:ladder-seeds}). The distance to the corpus rises with each added
direction and falls on release. Support drops to $0$--$1$ windows by the clearance
stage on jog and backward walk.

\begin{table}[t]
\centering
\caption{Steering vs.\ matched joint offsets at $a{=}2$ over three
randomized seeds (mean $\pm$ SD; brackets: deterministic run). The
$22$ triples cover five bases $\times$ five directions, minus three steered
clearance runs that fell deterministically. Falls are per seed ($0$, $1$, $2$), out of
$22$. Speed is the median ratio to base
of arm directions on walk, jog and backward walk, drift the median
$|$heading change$|$ of non-turn directions, and turn the median heading
change of the turn direction, all over runs that stayed up. Pose is relative to the seed's steered run.}
\label{tab:forcing-seeds}
\scriptsize
\setlength{\tabcolsep}{3pt}
\begin{tabular}{@{}lccccc@{}}
\toprule
Interface & Falls & Pose & Speed & Drift ($^\circ$) & Turn ($^\circ$) \\
\midrule
Steering               & 0, 0, 0 [0] & $1.00$ & $0.95{\pm}0.02$ [0.90] & $16{\pm}3$ [20] & $295{\pm}10$ [279] \\
Offset on action       & 4, 5, 5 [5] & $1.37{\pm}0.03$ [1.38] & $0.98{\pm}0.10$ [0.89] & $36{\pm}8$ [29] & $192{\pm}4$ [177] \\
Offset on servo target & 1, 1, 2 [1] & $0.73{\pm}0.00$ [0.76] & $0.92{\pm}0.01$ [0.91] & $23{\pm}3$ [19] & $92{\pm}13$ [80] \\
\bottomrule
\end{tabular}
\end{table}

\AppFloatBarrier

\section{Training on LAFAN1}
\label{app:lafan1}

The released BFM-Zero model compared in
Appendix~\ref{app:composition-baselines} was trained on LAFAN1, and ours
on $129{,}785$ BONES-SEED clips. To separate the method from the data, we
trained our method on LAFAN1. On the same $40$ clips and under the same
termination rule, it completes $17$ whole clips and BFM-Zero none, each
in its own simulator. Over the first $15$\,s the counts are $34$ and
$19$. Steering still works for speed and turn, but the arm directions are
weaker than on BONES-SEED.

\paragraph{Setup.}
We use our own G1 retarget of LAFAN1 ($441{,}080$ frames at $50$\,Hz;
BFM-Zero's bank, from its own retarget, has $441{,}122$).
Six of the $40$ clips show a fall and a get-up.
The encoder and predictor are those of Section~\ref{sec:method}
($z\in\mathbb{R}^{64}$, five past frames of context, $H=10$), pretrained
on all $40$ clips for $3{,}000$ updates instead of $50{,}000$, since with four
clips held out the next-chunk loss was lowest at $2{,}000$--$3{,}000$
updates. The
tracker has the architecture and PPO recipe of our BONES-SEED tracker and
is trained from scratch, without the later fine-tuning stages. Resets
are drawn uniformly over clips and start frames. We
report the last checkpoint ($27.0$B frames) and an earlier one ($6.0$B).
Evaluation uses the same $40$ clips in Isaac Lab with the
Newton/MuJoCo-Warp backend of training, the mean action and no domain
randomization.

\paragraph{Evaluation settings.}
Table~\ref{tab:lafan1} uses three settings.
\begin{description}
\item[Terminating.] A run ends when the pelvis height error or an
  end-effector (ankle or wrist) height error exceeds $0.25$\,m, or the
  pelvis orientation error exceeds $1$\,rad. A whole clip succeeds if it
  reaches its end.
\item[No termination.] The setting of BFM-Zero's evaluation. Each clip
  runs to its end without reset. A (relative) fall is recorded when the
  pelvis stays more than $0.25$\,m below the reference pelvis for $0.5$\,s.
\item[Windows.] $20$ ten-second windows per clip at uniformly sampled
  start frames, scored with the terminating rule (\emph{complete}) and
  with a torso-height fall test (\emph{fall-free}).
\end{description}

\paragraph{Results.}
At $27.0$B, $17$ of the $40$ clips complete under termination, up from
$10$ at $6.0$B (Table~\ref{tab:lafan1}).

\begin{table}[t]
\centering
\caption{Our method trained and evaluated on the 40 LAFAN1 clips. MPJPE in
mm, mean over clips with the median in parentheses. No-termination rows
include steps after a fall. Window rows average the pre-fall steps.
$^\dagger$Successful clips only, frame-weighted. $^\ddagger$Torso-height
fall test: torso below $0.4$\,m while the reference pelvis is at or above
$0.4$\,m. With the relative fall test, $688/800$ windows are fall-free at
$27.0$B.}
\label{tab:lafan1}
\small
\setlength{\tabcolsep}{3.5pt}
\begin{tabular}{@{}llccc@{}}
\toprule
Setting & Frames & Success & MPJPE-L & MPJPE-G \\
\midrule
Terminating, whole clip         & 6.0B  & 10/40 & $22.5^\dagger$ & $85.2^\dagger$ \\
                                & 27.0B & 17/40 & $22.1^\dagger$ & $77.7^\dagger$ \\
\addlinespace
No termination, whole clip      & 6.0B  & 23/40 fall-free & 110.9 (42.7) & 829.9 (398.4) \\
                                & 27.0B & 24/40 fall-free & 98.7 (27.8) & 578.3 (142.0) \\
\addlinespace
$20\times10$\,s windows per clip & 6.0B  & 665/800 fall-free$^\ddagger$ & 41.0 (31.0) & 157.1 (127.6) \\
                                & 6.0B  & 594/800 complete & -- & -- \\
                                & 27.0B & 668/800 fall-free$^\ddagger$ & 32.5 (25.5) & 101.6 (78.6) \\
                                & 27.0B & 639/800 complete & -- & -- \\
\bottomrule
\end{tabular}
\end{table}

\paragraph{Side by side with BFM-Zero.}
Table~\ref{tab:lafan1-bfm} compares the two models; its terminating rows
use the same three termination terms and thresholds for both. The
comparison is not controlled:
BFM-Zero runs in its MuJoCo deployment simulator on its own retarget, and
ours in Isaac Lab on ours. Both evaluations compute MPJPE-L after
subtracting only the pelvis position, so heading drift raises MPJPE-L as
well as MPJPE-G. Most of BFM-Zero's terminations are orientation
failures, which its evaluation attributes to heading drift. In a separate
run without termination, $23$ clips first failed on orientation,
and $22$ of them had a tilt error below $0.2$\,rad at failure.
Without termination, BFM-Zero falls less often than ours, but its MPJPE-L
on clips without a fall is $196.4$\,mm against $23.4$\,mm.

\begin{table}[t]
\centering
\caption{BFM-Zero (released checkpoint, its MuJoCo deployment simulator
and retarget) and our 27.0B LAFAN1 model (Isaac Lab) on the 40 LAFAN1
clips. MPJPE in mm, frame-weighted over the scored steps of the stated
clips. The torso-height rows exclude every clip in which the torso
reaches the floor: five fall-and-get-up clips for BFM-Zero, all six for
ours. $^*$From a separate run in which a fall ends the clip.}
\label{tab:lafan1-bfm}
\small
\setlength{\tabcolsep}{3.5pt}
\begin{tabular}{@{}lcc@{}}
\toprule
 & BFM-Zero & Ours, 27.0B \\
\midrule
\multicolumn{3}{@{}l}{\emph{Whole clip, terminating}} \\
Success                                      & 0/40 & 17/40 \\
Mean steps before termination                & $1{,}028$ & $7{,}904$ \\
First failure: orientation / end effector / pelvis height & 24 / 15 / 1 & 2 / 19 / 2 \\
MPJPE-L / G, successful clips                & -- & 22.1 / 77.7 \\
Fall-and-get-up clips completed              & 0/6 & 0/6 \\
\addlinespace
\multicolumn{3}{@{}l}{\emph{Whole clip, no termination}} \\
MPJPE-L / G, all clips                       & 211.1 / $6{,}372$ & 85.0 / 501.9 \\
Clips without a relative fall                & 30/40 & 24/40 \\
\quad excluding the six fall-and-get-up clips & 30/34 & 24/34 \\
MPJPE-L / G, clips without a fall            & 196.4 / $6{,}601$ & 23.4 / 89.4 \\
MPJPE-L / G, fall-and-get-up clips (all fall) & 295.7 / $3{,}777$ & 353.0 / $2{,}259$ \\
MPJPE-L / G, pre-fall prefix, all clips      & 201.7 / $6{,}462$ & 25.8 / 108.1$^*$ \\
\addlinespace
\multicolumn{3}{@{}l}{\emph{First 15\,s, terminating}} \\
Success                                      & 19/40 & 34/40 \\
MPJPE-L / G, successful clips                & 42.5 / 298 & 19.6 / 52.2 \\
\addlinespace
\multicolumn{3}{@{}l}{\emph{First 15\,s, no termination}} \\
MPJPE-L / G, all clips                       & 72.0 / 567 & 35.7 / 183.7 \\
Torso never below $0.4$\,m                   & 35/40 & 34/40 \\
MPJPE-L / G, those clips                     & 62.7 / 513 & 19.5 / 78.7 \\
\bottomrule
\end{tabular}
\end{table}

\AppFloatBarrier

\end{document}